%% file: main.tex
\documentclass{article}

\usepackage[final]{nips_2018}

\usepackage{nips_2018}

\usepackage[utf8]{inputenc} 
\usepackage[T1]{fontenc}    
\usepackage{hyperref}       
\usepackage{url}            
\usepackage{booktabs}       
\usepackage{amsfonts}       
\usepackage{nicefrac}       
\usepackage{microtype}      
\usepackage{color}

\usepackage{qiangstyle}
\usepackage{setspace}

\usepackage{graphicx}
\usepackage{tabu}

\title{Variational Inference with Tail-adaptive \emph{f}-Divergence}

\author{
{\rm Dilin Wang}\\
{\rm UT Austin}\\
{\rm \texttt{dilin@cs.utexas.edu}} 
\and
{\rm Hao Liu \thanks{Work done at UT Austin}}\\
{\rm UESTC}\\
{\rm \texttt{uestcliuhao@gmail.com}} 
\and
{\rm Qiang Liu}\\
{\rm UT Austin}\\
{\rm \texttt{lqiang@cs.utexas.edu}} 
}

\begin{document}

\input{tex/main_include.tex}

\input{tex/appendix.tex}\textbf{}

\end{document}

%% file: tex/main_include.tex
\maketitle


\begin{abstract}
Variational inference with $\alpha$-divergences 
has been widely used in modern probabilistic machine learning.
Compared to Kullback-Leibler (KL) divergence, 
a major advantage of using $\alpha$-divergences (with positive $\alpha$ values) 
is their 
\emph{mass-covering} property. 
However, 
estimating and optimizing $\alpha$-divergences require 
to use importance sampling, 
which may have 
 large or infinite variance due to 
 heavy tails of importance weights. 
In this paper, we propose a new class of \emph{tail-adaptive} $f$-divergences 
that adaptively change 
the convex function $f$  
with the tail distribution of the importance weights,   
in a way that theoretically guarantees finite moments, while simultaneously achieving mass-covering properties. 
We test our method on Bayesian neural networks, and apply it to improve a recent soft actor-critic (SAC) algorithm \citep{haarnoja2018soft} in deep reinforcement learning. 
Our results show that our approach yields significant advantages compared with existing methods based on classical KL and  $\alpha$-divergences. 
\end{abstract}

\input{tex/intro}
\input{tex/f_divergence}

\input{tex/experiments}

\section{Conclusion}
\label{sec:con}
In this paper, we present a new class of tail-adaptive $f$-divergence and exploit its application in variational inference and reinforcement learning. 
Compared to classic $\alpha$-divergence, our
approach  guarantees finite moments of 
the density ratio and provides more stable importance weights 
and gradient estimates. 
Empirical results on Bayesian neural networks and 
reinforcement learning indicate that our approach outperforms
standard $\alpha$-divergence, 
 especially for high dimensional multi-modal distribution.

\section*{Acknowledgement}
This work is supported in part by NSF CRII 1830161.
We would like to acknowledge Google Cloud for their support.

\bibliography{nips}
\bibliographystyle{nips2018}

%% file: tex/intro.tex

\section{Introduction}
Variational inference (VI) \citep[e.g.,][]{jordan1999introduction, wainwright2008graphical} 
has been established as a powerful tool in modern probabilistic machine learning  
for approximating intractable posterior distributions. 
The basic idea is to turn the approximation problem into an optimization problem,  
which finds the best approximation of an intractable distribution from a family of  tractable distributions 
by minimizing a divergence objective function.  
Compared with Markov chain Monte Carlo (MCMC), which is known to be consistent 
but suffers from slow convergence, VI provides biased results but is often practically faster.   
Combined with techniques like stochastic optimization \citep{ranganath2014black, hoffman2013stochastic} and reparameterization trick \citep{kingma2013auto}, 
VI has become a major technical approach for advancing Bayesian deep learning, deep generative models and deep reinforcement learning  \citep[e.g.,][]{kingma2013auto, gal2016dropout, levine2018reinforcement}.  

A key component of successful variational inference   lies on 
 choosing a proper divergence metric. 
Typically, closeness is defined by the KL divergence  $\KL(q ~||~ p)$ \citep[e.g.,][]{jordan1999introduction}, 
where $p$ is the intractable distribution of interest and $q$ is a simpler distribution constructed to  approximate $p$.  
However, VI with KL divergence 
often under-estimates the variance and may miss  important local modes of the true posterior  \citep[e.g.,][]{christopher2016pattern, blei2017variational}. 
To mitigate this issue, 
alternative metrics have been studied in the literature, a large portion of which are special cases of  $f$-divergence \citep[e.g.,][]{CIT-004}: 
\begin{align} \label{fo}
D_f (p~||~q) = \E_{x\sim q} \left [ f\left (\frac{p(x)}{q(x)} \right ) - f(1) \right ], 
\end{align}
where $f\colon \R_+ \to \R $ is any convex function.  
The most notable class of $f$-divergence that has been exploited in VI is $\alpha$-divergence, which takes $f(t)=t^\alpha/(\alpha(\alpha-1))$ 
for $\alpha\in \R\setminus\{0,1\}$. 
By choosing different $\alpha$, we get a large number of well-known divergences as special cases, 
including the standard KL divergence objective  $\KL(q~||~p)$ ($\alpha\to 0$),  
the KL divergence with the reverse direction $\KL(p~||~q)$ ($\alpha\to 1$) 
and the $\chi^2$ divergence ($\alpha=2$). In particular, 
the use of general $\alpha$-divergence in VI has been widely discussed  \citep[e.g.,][]{minka2005divergence, hernandez2016black, li2016renyi};  
the reverse KL divergence is used in expectation  propagation \citep{minka2001expectation,opper2005expectation}, importance weighted auto-encoders \citep{burda2015importance}, 
and the cross entropy method \citep{de2005tutorial};
$\chi^2$-divergence is exploited for VI \citep[e.g.,][]{dieng2017variational}, but is more extensively studied in the context of adaptive importance sampling (IS) \citep[e.g.,][]{cappe2008adaptive, ryu2014adaptive, cotter2015parallel}, since it coincides with the variance of the IS estimator with $q$ as the proposal. 

A major motivation of using $\alpha$-divergence 
contributes to its \emph{mass-covering} property: 
when $\alpha > 0$, the optimal approximation $q$ tends to cover more modes of $p$, and hence better accounts for the uncertainty in $p$. 
Typically, larger values of $\alpha$ enforce  
stronger mass-covering properties.
In practice, however, $\alpha$ divergence and its gradient 
need to be estimated empirically using samples from $q$. 
Using large $\alpha$ values may cause  
high or infinite variance in the estimation because it involves estimating the $\alpha$-th power of the density ratio  $p(x)/q(x)$, which is likely distributed with a heavy or fat tail \citep[e.g.,][]{resnick2007heavy}. 
In fact, when $q$ is very different from $p$, the  expectation of 
ratio 
$(p(x)/q(x))^\alpha$ can be infinite 
(that is, $\alpha$-divergence does not exist).
This makes it problematic to use large $\alpha$ values, 
despite the mass-covering property it promises. 
In addition, it is reasonable to expect that  
the optimal setting of $\alpha$ should vary across training processes and learning tasks. 
Therefore, it is desirable to 
design an approach to choose $\alpha$ \emph{adaptively} and \emph{automatically}
as $q$ changes during the training iterations, 
according to the distribution of the ratio $p(x)/q(x)$. 

Based on theoretical observations on $f$-divergence and fat-tailed distributions,  
we design a new class of $f$-divergence  
which is \emph{tail-adaptive} in that it uses different $f$ functions according to the tail distribution of the density ratio $p(x)/q(x)$ to simultaneously obtain stable empirical estimation and a strongest possible  mass-covering property.  
This allows us to derive a new adaptive $f$-divergence-based variational inference by combining it with stochastic optimization and reparameterization gradient estimates. 
Our main method (Algorithm~\ref{alg1}) has a simple  form, 
which replaces the $f$ function in \eqref{fo} with a rank-based function of the empirical density ratio $w=p(x)/q(x)$ 
at each gradient descent step of $q$, 
whose variation depends on the distribution of $w$ and does not explode regardless the tail of $w$.  

Empirically, we show that our method can better recover multiple modes for variational inference. In addition, we apply our method to improve a recent soft actor-critic (SAC) algorithm \citep{haarnoja2018soft} in reinforcement learning (RL), showing that our method can be used to optimize multi-modal loss functions in RL more efficiently. 

%% file: tex/f_divergence.tex
\section{$f$-Divergence and Friends} 
\label{sec:f_div}
Given a distribution $p(x)$ of interest, 
we want to approximate it with a simpler distribution
from a family $\{q_\theta(x) \colon \theta \in \Theta\}$, where $\theta$ is the variational
parameter that we want to optimize. 
We approach this problem by minimizing the $f$-divergence between $q_\theta$ and $p$: 
\begin{align}\label{eq:Lq}
\min_{\theta\in \Theta} 
\left \{ D_f (p~||~q_\theta) =  \E_{x\sim q_\theta}\left [f\left (\frac{p(x)}{q_\theta(x)}\right ) - f(1) \right ],   
\right \}
\end{align}
where $f \colon \R_{+} \to \R$ is any twice differentiable convex function. 
It can be shown by Jensen's inequality that $\D_f(p~||~q)\geq 0$ for any $p$ and $q$. 
Further, if $f(t)$ is strictly convex at $t = 1$, 
then $D_f(p~||~q) = 0$ implies $p= q$. 
The optimization in \eqref{eq:Lq} can be solved approximately 
using stochastic optimization in practice by 
approximating the expectation $\E_{x\sim q_\theta}[\cdot]$ using 
 samples drawing from $q_\theta$ at each iteration. 

The $f$-divergence includes a large spectrum of important divergence measures. It includes KL divergence in both directions,  
\begin{align}
    \KL(q~||~p) = \E_{x\sim q}\left [\log\frac{ q(x)}{p(x)}\right], 
    &&
    \KL(p~||~q) = \E_{x\sim q} \left [\frac{p(x)}{q(x)}\log\frac{p(x)}{q(x)}\right ], 
\end{align}
which correspond to $f(t)=-\log t$ and $f(t) =  t \log t$, respectively. 
$\KL(q~||~p)$ is the typical objective function used in variational inference; 
the reversed direction $\KL(p~||~q)$
is also  used in various settings \citep[e.g.,][]{minka2001expectation, opper2005expectation, de2005tutorial, burda2015importance}.

More generally, $f$-divergence includes the class of $\alpha$-divergence,  
which takes  $f_\alpha(t) = t^\alpha/(\alpha(\alpha-1))$, $\alpha \in \R \setminus \{0,1\}$ and hence 
\begin{align}\label{eq:alpha}
    D_{f_\alpha}(p~||~q) = \frac{1}{\alpha (\alpha - 1)}\E_{x\sim q} \left [ \left ( \frac{p(x)}{q(x)} \right )^\alpha -1 \right ]. 
\end{align}
One can show that $\KL(q~||~p)$ and $\KL(p~||~q)$ 
are the limits of $D_{f_\alpha}(q~||~p)$ when $\alpha \to 0$ and $\alpha \to 1$, respectively. Further, 
one obtain Helinger distance and $\chi^2$-divergence as $\alpha =1/2$ and $\alpha =2$, respectively. 
In particular, $\chi^2$-divergence ($\alpha=2$) plays an important role in adaptive importance sampling, 
because it equals the variance of the importance weight $w=p(x)/q(x)$ and minimizing $\chi^2$-divergence corresponds to finding an optimal importance sampling proposal.  

\section{$\alpha$-Divergence and Fat Tails}  
\label{sec:alpha_heavy}

A major motivation of using $\alpha$ divergences 
as the objective function for approximate inference is their \emph{mass-covering} property (also known as the zero-avoiding behavior). 
This is because $\alpha$-divergence is proportional to 
the $\alpha$-th moment of the density ratio $p(x)/q(x)$. 
When $\alpha$ is positive and large, 
large values of $p(x)/q(x)$ are strongly penalized,  preventing the case of $q(x)\ll p(x)$. 
In fact, whenever $D_{f_\alpha}(p~||~q)<\infty$, 
we  have $p(x) > 0$ imply $q(x) > 0$. 
This means that the probability mass and local modes of $p$ are taken into account in $q$ properly.   


Note that the case when $\alpha \leq 0$ exhibits the opposite property, that is, $p(x)=0$ must imply $q(x)=0$ to make $D_{f_\alpha}(q||p)$ finite when $\alpha \leq 0$; this includes the typical KL divergence $\KL(q~||~p)$ ($\alpha = 0$), which is often criticized for its tendency to under-estimate the uncertainty.

Typically, 
using larger values of $\alpha$ enforces stronger \emph{mass-covering} properties. 
In practice, however, larger values of $\alpha$ also increase the variance of the empirical estimators, making it highly challenging to optimize. 
In fact, the expectation in \eqref{eq:alpha} may not even exist when $\alpha$ is too large. This is because the density ratio $w:=p(x)/q(x)$ often has a fat-tailed distribution.  


A non-negative random variable $w$ is called fat-tailed\footnote{Fat-tailed distributions is a sub-class of heavy-tailed distributions, which are distributions whose tail probabilities decay slower than exponential functions, that is, $\lim_{t\to+\infty}\exp(\lambda t) \bar F_w(t)=\infty$ for all $\lambda > 0$.} \citep[e.g.,][]{resnick2007heavy} if its tail probability $\bar F_w(t):= \prob(w \geq t)$ 
is asymptotically equivalent to $t^{-\alpha_*}$ as $t\to +\infty$ for some finite positive number $\alpha_*$ (denoted by $\bar F_w(t) \sim t^{-\alpha_*}$), which means that 
$$\bar F_w(t) = t^{-\alpha_*} L(t),$$ where $L$ is a slowly varying function that satisfies $\lim_{t\to +\infty}L(ct)/L(t) = 1$ for any $c> 0$. 
Here $\alpha_*$ determines the fatness of the tail and is called the tail index of $w$. 
For a fat-tailed distribution with index $\alpha_*$, 
its $\alpha$-th moment exists only if $\alpha < \alpha_*$, that is, $\E[w^\alpha] <\infty $  iff $\alpha < \alpha_*$. 
It turns out the density ratio $w:=p(x)/q(x)$, when $x\sim q$, tends to have a fat-tailed distribution when $q$ is more peaked than $p$.  
The example below illustrates this with simple Gaussian distributions.

\begin{exa}
\label{ex:ex1}
Assume $p(x) = \normal(x;0, \sigma_p^2)$ and $q(x) = \normal(x; 0, \sigma_q^2)$. 
Let $x\sim q$ and $w = p(x)/q(x)$ the density ratio. If $\sigma_p > \sigma_q$, then $w$ has a fat-tailed distribution with index $\alpha_* = \sigma^2_p/(\sigma_p^2 - \sigma_q^2).$ 

On the other hand, if $\sigma_p \leq \sigma_q$, then $w$ is bounded and not fat-tailed (effectively, $\alpha_* = +\infty$).  
\end{exa}

By the definition above, if the importance weight $w = p(x)/q(x)$ has a tail index $\alpha_*$, the $\alpha$-divergence $D_{f_\alpha}(p~||~q)$ exists only if $\alpha < \alpha_*$. 
Although it is desirable to use $\alpha$-divergence with large values of $\alpha$ as VI objective function, 
it is important to keep $\alpha$ smaller than $\alpha_*$ to ensure that the objective and gradient are well defined. 
The problem, however, is that the tail index $\alpha_*$ is unknown in practice, 
and may change dramatically (e.g., even from finite to infinite) as $q$ is updated during the optimization process.
This makes it suboptimal to use a pre-fixed $\alpha$  value. 
One potential way to address this problem is to estimate the tail index $\alpha^*$ empirically 
at each iteration using a tail index estimator  \citep[e.g.,][]{hill1975simple, vehtari2015pareto}. 
Unfortunately, 
tail index estimation is often challenging and requires a large number of samples. 
The algorithm may become unstable if $\alpha_*$ is over-estimated. 


\section{Hessian-based Representation of $f$-Divergence} 
\label{sec:f_grad}
In this work, we address the aforementioned problem by designing 
a generalization of $f$-divergence in which $f$ adaptively changes with $p$ and $q$, in a way that always guarantees the existence of the expectation,  while simultaneous achieving (theoretically) strong mass-covering equivalent to that of the $\alpha$-divergence with  $\alpha = \alpha^*$.   

One challenge of designing such adaptive $f$  
is that the convex constraint over function $f$ is difficult to express computationally. 
Our first key observation is that it is easier to specify a convex function $f$ through its second order derivative $f''$, which can be any non-negative function. It turns out $f$-divergence, as well as its gradient, can be conveniently expressed using $f''$, without explicitly defining the original  $f$.  
\begin{pro}
1) Any twice differentiable convex function $f\colon \R_+ \cup\{0\} \to \R$ with finite $f(0)$ can be decomposed into linear and nonlinear components as follows  
\begin{align} \label{fff}
f(t) = (a t + b) ~+~\int_0^\infty (t - \mu)_+ h(\mu) d\mu, 
\end{align}
where $h$ is a non-negative function, $(t)_+  = \max(0,t)$, 
and $a$,$b\in \R$. 
In this case, $h = f''(t),$ $a=f'(0)$ and $b=f(0)$.  
Conversely, any non-negative function $h$ and $a,b\in \R$ specifies a convex function. 

2) This allows us to derive an alternative representation of $f$-divergence: 
\begin{align}
D_f(p~||~q) = \int_0^\infty f''(\mu) \E_{x\sim q} \left [  \left (\frac{p(x)}{q(x)} - \mu \right)_+  \right ] d\mu  ~ - ~ c, 
\label{eq:f_sec_dev}
\end{align}
where $c := \int_0^1 f''(\mu)(1-\mu)d\mu =  f(1)-f(0)-f'(0)$ is a constant. 
\end{pro}
\begin{proof}
If $f(t) = (at+b) + \int_0^\infty (t-\mu)_+ h(\mu)d\mu$, calculation shows 
$$
f'(t) = a + \int_0^t h(\mu) d\mu, ~~~f''(t) = h(t).
$$
Therefore, $f$ is convex iff $h$ is non-negative. See Appendix for the complete proof. 
\end{proof}
%
Eq~(\ref{eq:f_sec_dev}) suggests that all $f$-divergences 
are conical combinations of a set of special $f$-divergences of form $\E_{x\sim q}  [  ({p(x)}/{q(x)} - \mu  )_+  -f(1)]$ with $f(t) = (t-\mu)_+$.   
Also, every $f$-divergence is completely specified by the Hessian $f''$, 
meaning that adding $f$ with any linear function $at +b$ does not change $D_f(p~||~q)$. 
%
Such integral representation of 
$f$-divergence 
is not new; see e.g.,  \citet[][]{feldman1989note, osterreicher2003f, liese2006divergences, reid2011information, sason2018f}.

For the purpose of minimizing $\D_f(p~||~q_\theta)$ ($\theta\in \Theta$) in variational inference, we are more concerned with calculating the gradient, 
rather than the $f$-divergence itself. 
It turns out the gradient of $\D_f(p~||~q_\theta)$ is also directly related to Hessian $f''$ in a simple way.  
\begin{pro}
1) Assume $\log q_\theta(x)$ is differentiable w.r.t. $\theta$, 
and $f$ is a differentiable convex function. 
For $f$-divergence defined in \eqref{eq:Lq}, we have 
\begin{align}\label{score}
\nabla_\theta D_f(p ~||~ q_\theta) = - \E_{x\sim q_\theta} \left [ \rho_f \left ( \frac{p(x)}{q_\theta(x)}\right) \nabla_\theta \log q_\theta(x) \right ],
\end{align}
where $\rho_f(t) =   f'(t) t - f(t)$ (equivalently, $\rho_f'(t) = f''(t)t$ if $f$ is twice differentiable).  

2) Assume $ x \sim q_\theta $ is generated by $x = g_\theta(\xi)$ where $\xi \sim q_0$ is a random seed and $g_\theta$ is a function that is differentiable w.r.t. $\theta$. Assume $f$ is twice differentiable and $\nabla_x\log(p(x)/q_\theta(x))$ exists. We have 
\begin{align}
\label{eq:repa}
\begin{split}
 & \nabla_\theta D_f(p ~||~ q_\theta) = - \E_{x=g_\theta(\xi), \xi\sim q_0}  \left [ 
\omeganewnew_f \left ( \frac{p(x)}{q_\theta(x)}\right)    \nabla_\theta g_\theta(\xi)  \nabla_x \log (p(x)/ q_\theta(x)) 
\right], 
\end{split}
\end{align}
where $\omeganewnew_f (t) = \rho_f'(t)t =  f''(t) t^2.$ 
\label{pro:gradient}
\end{pro}
The result above shows that the gradient of $f$-divergence depends on $f$ through $\rho_f$ or $\omeganewnew_f$. Taking $\alpha$-divergence $(\alpha\notin\{0,1\})$ as example, we have 
\begin{align*}
f(t)=t^\alpha/(\alpha(\alpha-1)), &&
\rho_f(t) = t^\alpha/\alpha, &&
\omeganewnew_f(t) = t^\alpha, 
\end{align*}
all of which are proportional to the power function $t^\alpha$. 
For $\KL(q~||~p)$, we have $f(t)=-\log t$, yielding $\rho_f(t) = \log t-1$ and  $\omeganewnew_f(t) =  1$;
for $\KL(p~||~q)$, we have $f(t) = t \log t$, yielding $\rho_f (t)=t$ and $\omeganewnew_f(t)= t$. 

The formulas in \eqref{score} and \eqref{eq:repa} 
are called the \emph{score-function gradient} and \emph{reparameterization gradient}~\citep{kingma2013auto}, respectively. 
Both equal the gradient in expectation, but are computationally different and yield empirical estimators with different variances. 
In particular, the score-function gradient in \eqref{score} is ``gradient-free'' in that it does not require calculating the gradient of the distribution $p(x)$ of interest, while \eqref{eq:repa} is ``gradient-based''
in that it involves $\nabla_x \log p(x).$
It has been shown that optimizing with reparameterization gradients tend to give   better empirical results because it leverages the gradient information $\nabla_x \log p(x)$, and yields a lower variance estimator for the gradient \citep[e.g.,][]{kingma2013auto}.



Our key observation is that we can directly specify $f$ through any increasing function $\rho_f$, or non-negative function $\omeganewnew_f$ in the gradient estimators, without explicitly defining $f$. 
\begin{pro} \label{pro:fffg}
Assume $f \colon \R_+ \to \R$ is convex and twice differentiable, then 

1) $\rho_f$  in \eqref{score} is a monotonically increasing function on $\R_+$. 
In addition, for any differentiable increasing function $\rho$, there exists a convex function $f$ such that $\rho_f  = \rho$; 

2) $\omeganewnew_f$  in \eqref{eq:repa} is non-negative on $\RR_+$, that is, $\omeganewnew_f(t)\geq 0$, $\forall t\in\RR_+$. In addition, for any non-negative function  $\omeganewnew$, there exists a convex function $f$ such that $\omeganewnew_f = \omeganewnew$; 

3) if $\rho_f'(t)$ is strictly increasing at $t=1$ (i.e., $\rho_f'(1) > 0$), or $\omeganewnew_f(t)$ is strictly positive at $t=1$ (i.e., $\omeganewnew_f(1) >0$), then $\D_f(p~||~q)=0$ implies $p=q$. 
\end{pro}
\begin{proof}
 Because $f$ is convex ($f''(t)\geq 0$), 
we have $\omeganewnew_f(t) = f''(t) t^2 \geq 0$ and  $\rho_f'(t) = f''(t)t\geq 0$ on $t\in \R_+$, that is, $\omeganewnew_f$ is non-negative and $\rho_f$ is increasing on $
\RR_+$.  
If $\rho_t$ is strictly increasing (or $\omeganewnew_f$ is strictly positive) at $t = 1$, we have $f$ is strictly convex at $t = 1$,  which guarantees $\D_f(p~||~q)=0$ imply $p=q$. 

 For non-negative function $\omeganewnew(t)$ (or increasing function $\rho(t)$) on $\RR_+$, any convex function $f$ whose second-order derivative equals $\omeganewnew(t)/t^2$ (or $\rho_f'(t)/t$) satisfies $\omeganewnew_f = \omeganewnew$ (resp. $\rho_f = \rho$). 
\end{proof}

\section{Safe $f$-Divergence with Inverse Tail Probability}
\label{our_approach}

The results above show that it is sufficient to find an increasing function $\rho_f$, or a non-negative function $\omeganewnew_f$ to obtain adaptive  $f$-divergence with computable gradients. %
%
In order to make the $f$-divergence ``safe'', 
we need to find $\rho_f$ or $\omeganewnew_f$ 
that adaptively depends on $p$ and $q$ such that the expectation in \eqref{score} and \eqref{eq:repa} always exists. 
Because the magnitude of $\nabla_\theta \log q_\theta(x)$, $\nabla_x \log (p(x)/q_\theta(x))$ and $\nabla_{\theta}g_\theta(\xi)$ are relatively small compared with the ratio $p(x)/q(x)$, 
we can mainly consider designing function $\rho$ (or $\omeganewnew$)  
such that they yield finite expectation $\E_{x\sim q}[\rho(p(x)/q(x))] <\infty$;  
meanwhile, we should also keep the function large, preferably with the same magnitude as $t^{\alpha_*}$, 
to provide a strong mode-covering property.  
As it turns out, the inverse of the tail probability naturally achieves all these goals. 
\begin{pro}
For any random variable $w$ with tail distribution $\bar F_w(t) := \Prob(w\geq t)$ and tail index $\alpha_*$, we have 
$$
\E[\bar F_w(w)^{\beta}] <\infty, ~~~~~~ \text{for any $\beta > -1$}.  
$$
Also, we have $\bar F_w(t)^{\beta} \sim t^{-\beta\alpha_*}$, and $\bar F_w(t)^\beta$ is always non-negative and monotonically increasing when $\beta < 0$.  \end{pro}
\begin{proof} 
Simply note that $
\E[\bar F_w(w)^{\beta}] 
= \int\bar F_w(t)^\beta d \bar F_\beta(t) = \int_{0}^1 t^{\beta} dt,$ 
which is finite only when $\beta >-1$. 
The non-negativity and monotonicity of $\bar F_w(t)^{\beta}$ are obvious. 
$\bar F_w(t)^{\beta} \sim t^{-\beta\alpha*}$ 
directly follows the definition of the tail index.  
\end{proof}

This motivates us to use $\bar F_w(t)^{\beta}$ to define $\rho_f$ or $\omeganewnew_f$, yielding two versions of  ``safe'' tail-adaptive $f$ divergences.
Note that here $f$ is defined implicitly through $\rho_f$ or $\omeganewnew_f$. Although it is possible to derive the corresponding $f$ and $D_f(p~||~q)$, there  is no computational need to do so, since optimizing the objective function only requires calculating the gradient, which is defined by $\rho_f$ or $\omeganewnew_f$.  
 
In practice, the explicit form of $\bar F_w(t)^\beta$ is unknown. We can approximate it based on empirical data drawn from $q$. Let $\{x_i\}$ be drawn from $q$ and $w_i = p(x_i)/q(x_i)$, then we can approximate the tail probability with $\hat{\bar F}_w(t) =\frac{1}{n}\sum_{i=1}^n \ind(w_i \geq t)$. 
Intuitively, this corresponds to assigning each data point a weight according to the rank of its
density ratio in the population. 
Substituting the empirical tail probability into the reparametrization gradient formula in \eqref{eq:repa} and running a gradient descent with stochastic approximation yields our main algorithm shown in Algorithm~\ref{alg1}.
The version with the score-function gradient is similar and is shown in Algorithm~2 in the Appendix. 
Both algorithms can be viewed as minimizing the implicitly constructed adaptive $f$-divergences, 
but correspond to using different $f$. 

Compared with typical VI with reparameterized gradients, our method assigns a weight $\rho_i = \hat{\bar F}_w(w_i)^\beta$, which is  proportional $\#w_i^{\beta}$ where $\#w_i$ denotes the rank of data $w_i$ in the population $\{w_i\}$. 
When taking $-1<\beta < 0$, this allows us to penalize places with high ratio $p(x)/q(x)$, but avoid to be overly aggressive.  
In practice, we find that simply taking $\beta = -1$ almost always yields the best empirical performance (despite needing $\beta>-1$ theoretically). 
By comparison, minimizing the classical $\alpha$-divergence would have a weight of $w_i^\alpha$; if $\alpha$ is too large, the weight of a single data point becomes dominant, making gradient estimate unstable.

\begin{algorithm}[t] 
\caption{Variational Inference with Tail-adaptive $f$-Divergence (with Reparameterization Gradient)}
\label{alg1} 
\begin{algorithmic} 
    \STATE Goal: Find the best approximation of $p(x)$ from $\{q_\theta \colon \theta\in \Theta\}$. Assume $x\sim q_\theta$ is generated by $x = g_\theta(\xi)$ where $\xi$ is a random sample from noise distribution $q_0$. 
    \STATE Initialize $\theta$, set an index $\beta$ (e.g., $\beta = -1$). 
    \FOR{iteration} 
    \STATE Draw $\{x_i \}_{i=1}^n \sim q_\theta $, generated by $x_i = g_\theta(\xi_i)$. 
    \STATE Let $w_i= p(x_i)/q_\theta(x_i)$, 
    $\hat{\bar F}_w(t) = \sum_{j=1}^n\ind(w_j \geq t)/n$, and 
    set $\omeganewnew_i = \hat{\bar F}_w(w_i)^{\beta}.$
    \STATE Update $\theta \gets \theta + \epsilon \Delta \theta$, with $\epsilon$ is step size, and 
 \begin{align*}
\Delta \theta  =\frac{1}{z_\omeganewnew}\sum_{i=1}^n\left [
\omeganewnew_i   \nabla_\theta g_\theta(\xi_i)  \nabla_x \log (p(x_i)/q_\theta(x_i))
\right],   ~~~\text{where}~~~z_\omeganewnew = \sum_{i=1}^n \omeganewnew_i.
 \end{align*}   
   \ENDFOR
\end{algorithmic}
\end{algorithm}

%% file: tex/experiments.tex
\section{Experiments}
\label{sec:exp}

In this section, we evaluate our adaptive $f$-divergence with different models.
We use reparameterization gradients as default since they have smaller variances
\citep{kingma2013auto} and normally yield better performance than 
score function gradients.
Our code is available at \url{https://github.com/dilinwang820/adaptive-f-divergence}.

\subsection{Gaussian Mixture Toy Example}
We first illustrate the approximation quality of 
our proposed adaptive $f$-divergence on Gaussian mixture models.
In this case, we set our target distribution to be a Gaussian mixture  
$p(x) =\sum_{i=1}^k \frac{1}{k}\normal(x; ~ \nu_i, 1)$,
for $x\in \RR^d$, 
where the elements of each mean vector $\nu_i$ is drawn from $\mathrm{uniform}([-s, s])$. 
Here $s$ can be viewed as controlling the Gaussianity of the target distribution: $p$ reduces to standard Gaussian distribution when $s=0$ and is increasingly multi-modal when $s$ increases. 
We fix the number of components to be $k=10$, and 
initialize the proposal distribution using 
$q(x)=\sum_{i=1}^{20} w_i\normal(x;~\mu_i, \sigma_i^2)$,
where $\sum_{i=1}^{20} w_i =1$. 

We evaluate the mode-seeking ability of how $q$ covers the modes of $p$ using a ``mode-shift distance'' $dist(p,q) := \sum_{i=1}^{10}  \min_j ||\nu_i - \mu_j||_2/10$, 
which is the average distance of each mode in $p$  
to its nearest mode in distribution $q$.  
The model is optimized using Adagrad with a constant learning rate $0.05$.
We use a minibatch of size 256 to approximate the gradient in each iteration. 
We train the model for $10,000$ iterations.
To learn the component weights,
we apply the Gumble-Softmax trick \citep{jang2016categorical, maddison2016concrete} with a temperature of $0.1$.
Figure~\ref{fig:gmm_toy} shows the result when we obtain random mixtures $p$ using $s=5$, 
when the dimension $d$ of $x$ equals $2$ and $10$,   
respectively. 
We can see that when the dimension is low ($=2$), all algorithms perform similarly well. However, as we increase the dimension to 10,
our approach with tail-adaptive $f$-divergence achieves the best performance.  



To examine the performance of variational approximation more closely,
we show in Figure~\ref{fig:gmm_scale} the average mode-shift distance
and the MSE of the estimated mean and variance as we gradually increase 
the non-Gaussianality of $p(x)$  by changing $s$. 
We fix the dimension to $10$.
We can see from Figure~\ref{fig:gmm_scale} that 
when $p$ is close to Gaussian (small $s$), all algorithms
perform well; when $p$ is highly non-Gaussian (large $s$), 
we find that our approach with adaptive weights 
significantly outperform other baselines.





\newcommand{\tmpht}{0.18}
\begin{figure*}[t]
\centering
\begin{tabular}{cccc}
(a) Mode-shift distance & (b) Mean & (c) Variance & \\
\raisebox{1.5em}{\rotatebox{90}{\scriptsize Avg. distance}}
\includegraphics[height=\tmpht\textwidth]{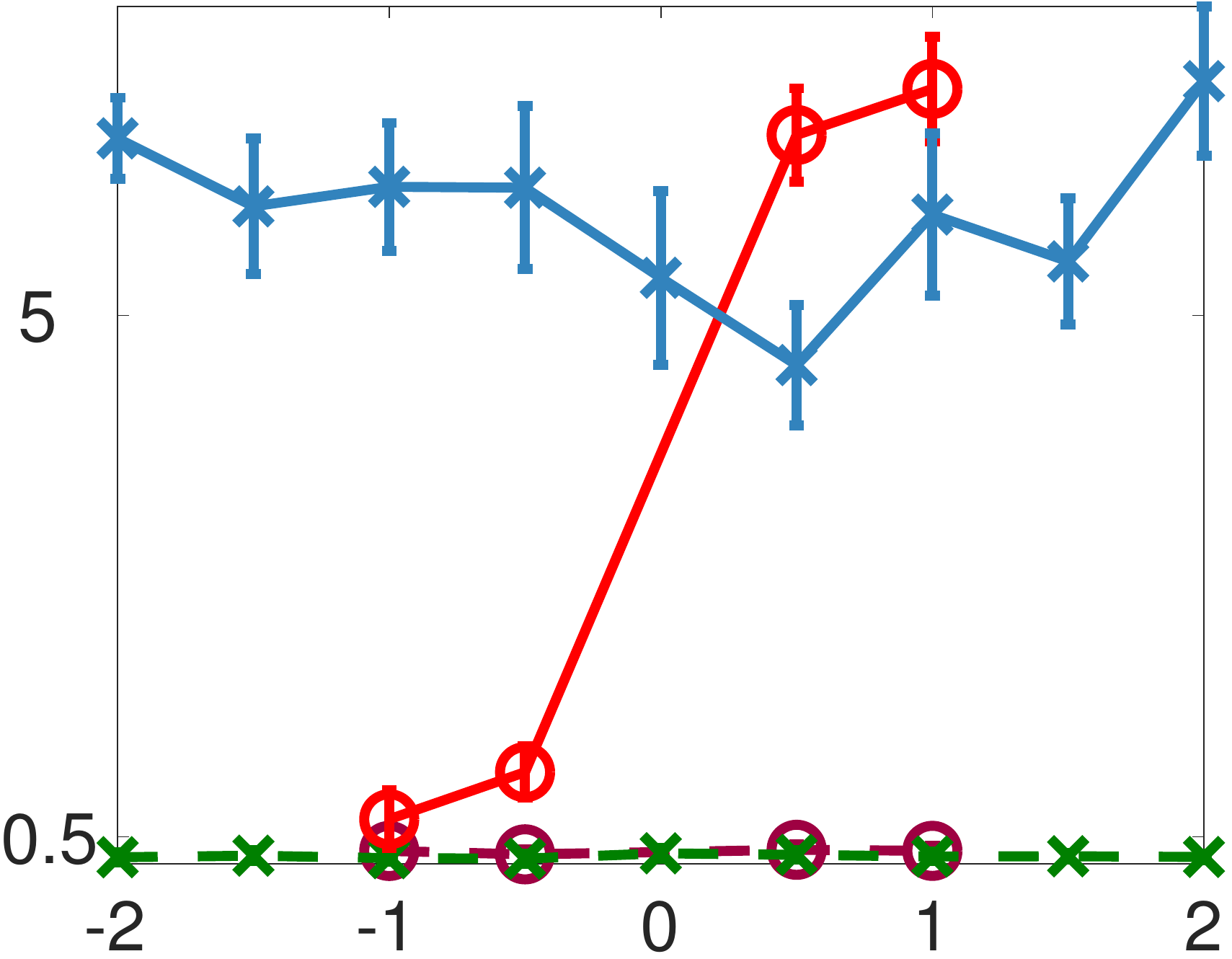} &
\raisebox{1.5em}{\rotatebox{90}{\scriptsize Log10 MSE}}
\includegraphics[height=\tmpht\textwidth]{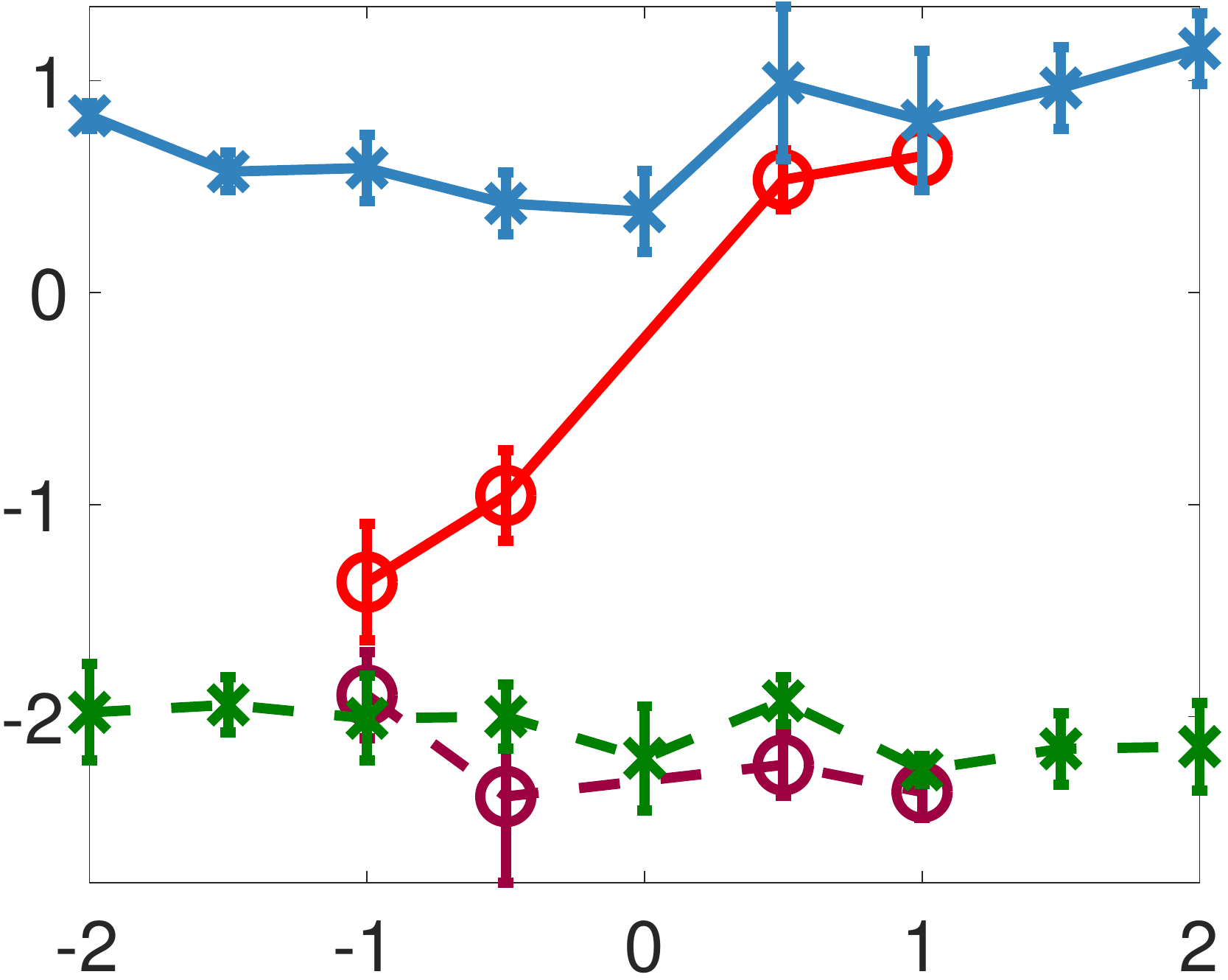} &
\raisebox{1.5em}{\rotatebox{90}{\scriptsize Log10 MSE}}
\includegraphics[height=\tmpht\textwidth]{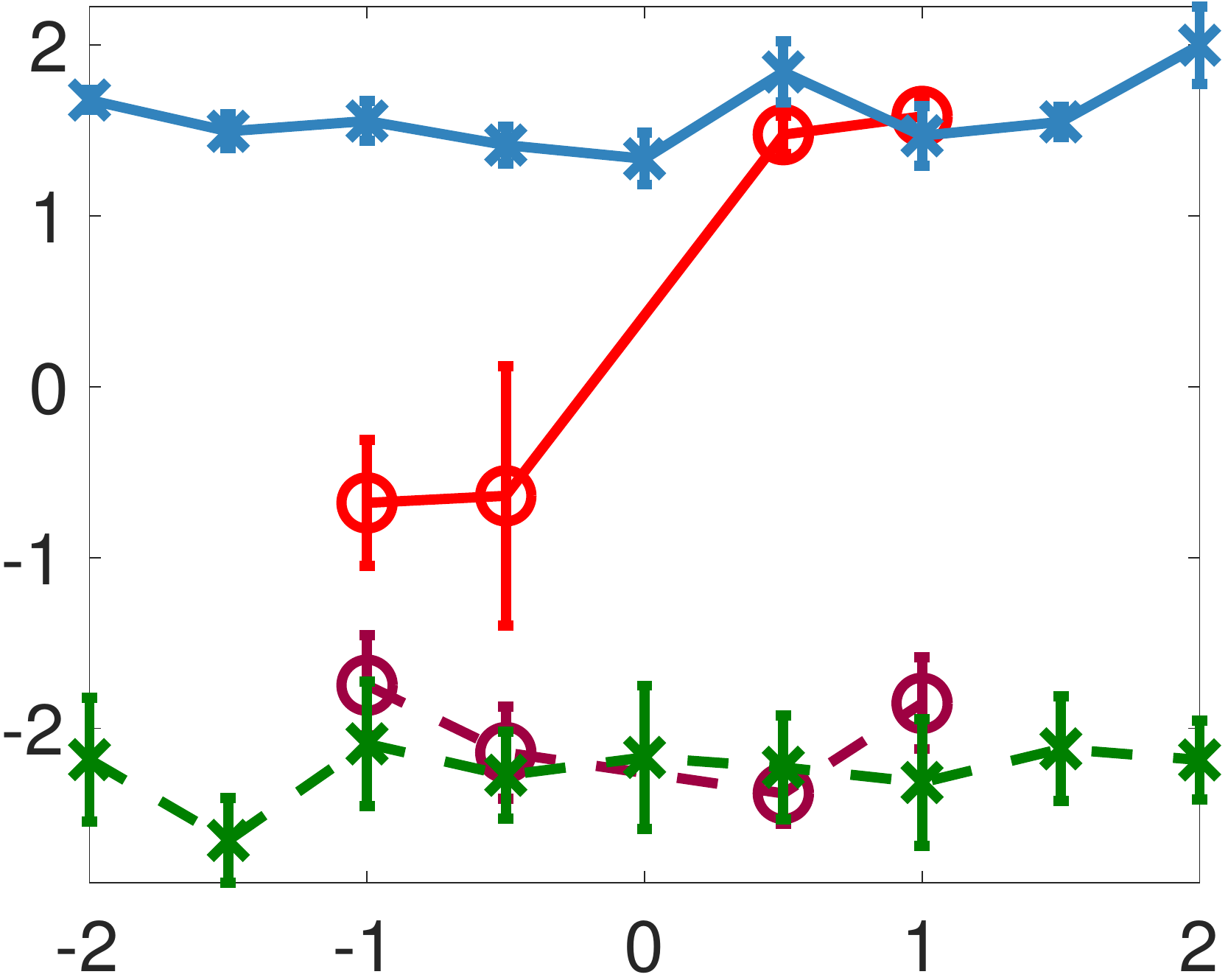}  & 
\raisebox{1.0em}{\includegraphics[width=0.15\textwidth]{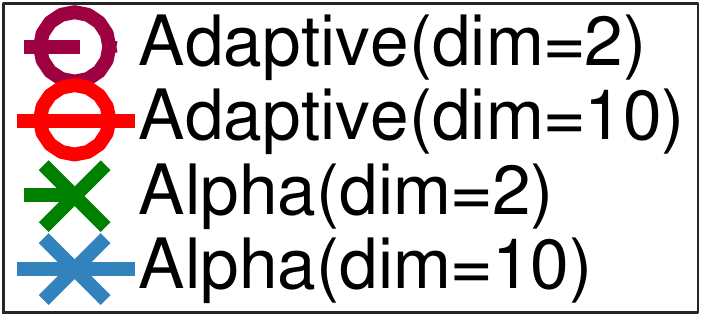}} \\
choice of $\alpha/\beta$ & choice of $\alpha/\beta$ &  choice of $\alpha/\beta$ & \\
\end{tabular}
\caption{\small (a) plots the mode-shift distance between $p$ and $q$; 
(b-c) show the MSE of mean and variance between the true posterior $p$ and our approximation $q$, respectively.
All results are averaged over 10 random trials.}
\label{fig:gmm_toy}
\end{figure*}

\begin{figure*}[t]
\centering
\begin{tabular}{cccc}
(a) Mode-shift distance & (b) Mean & (c) Variance  \\
\raisebox{1.5em}{\rotatebox{90}{\scriptsize Avg. distance}}
\includegraphics[height=\tmpht\textwidth]{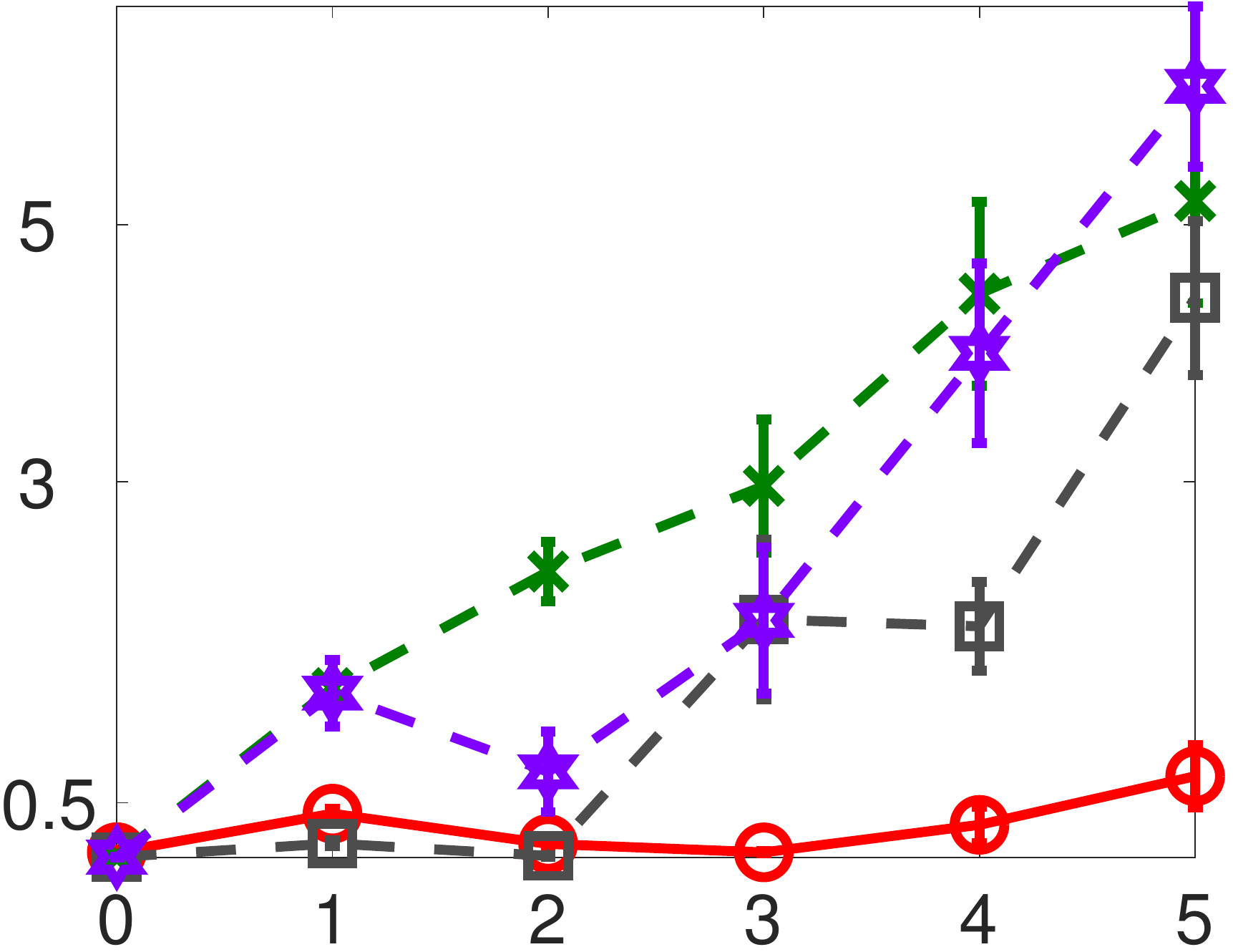} &
\raisebox{1.5em}{\rotatebox{90}{\scriptsize Log10 MSE}}
\includegraphics[height=\tmpht\textwidth]{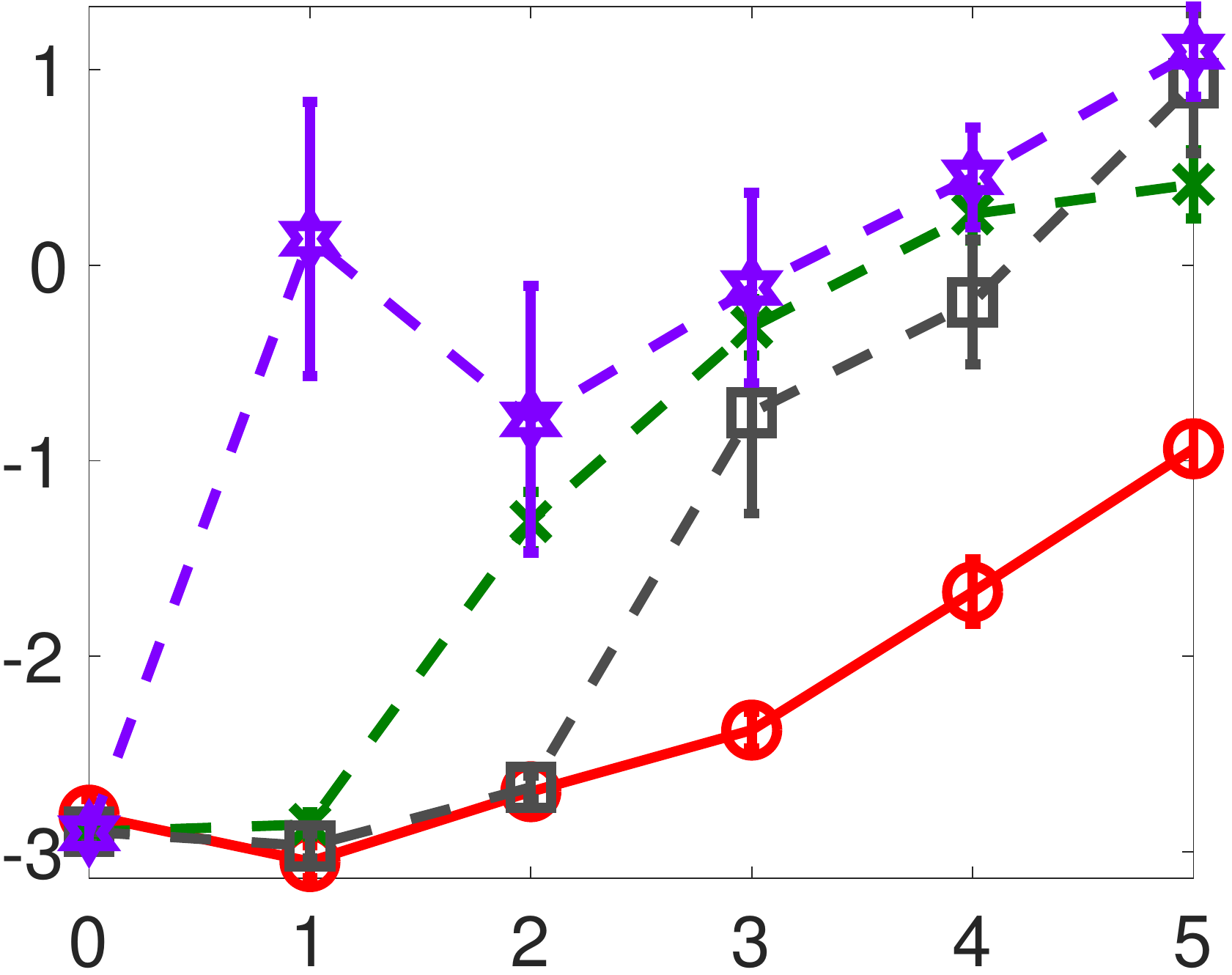} &
\raisebox{1.5em}{\rotatebox{90}{\scriptsize Log10 MSE}}
\includegraphics[height=\tmpht\textwidth]{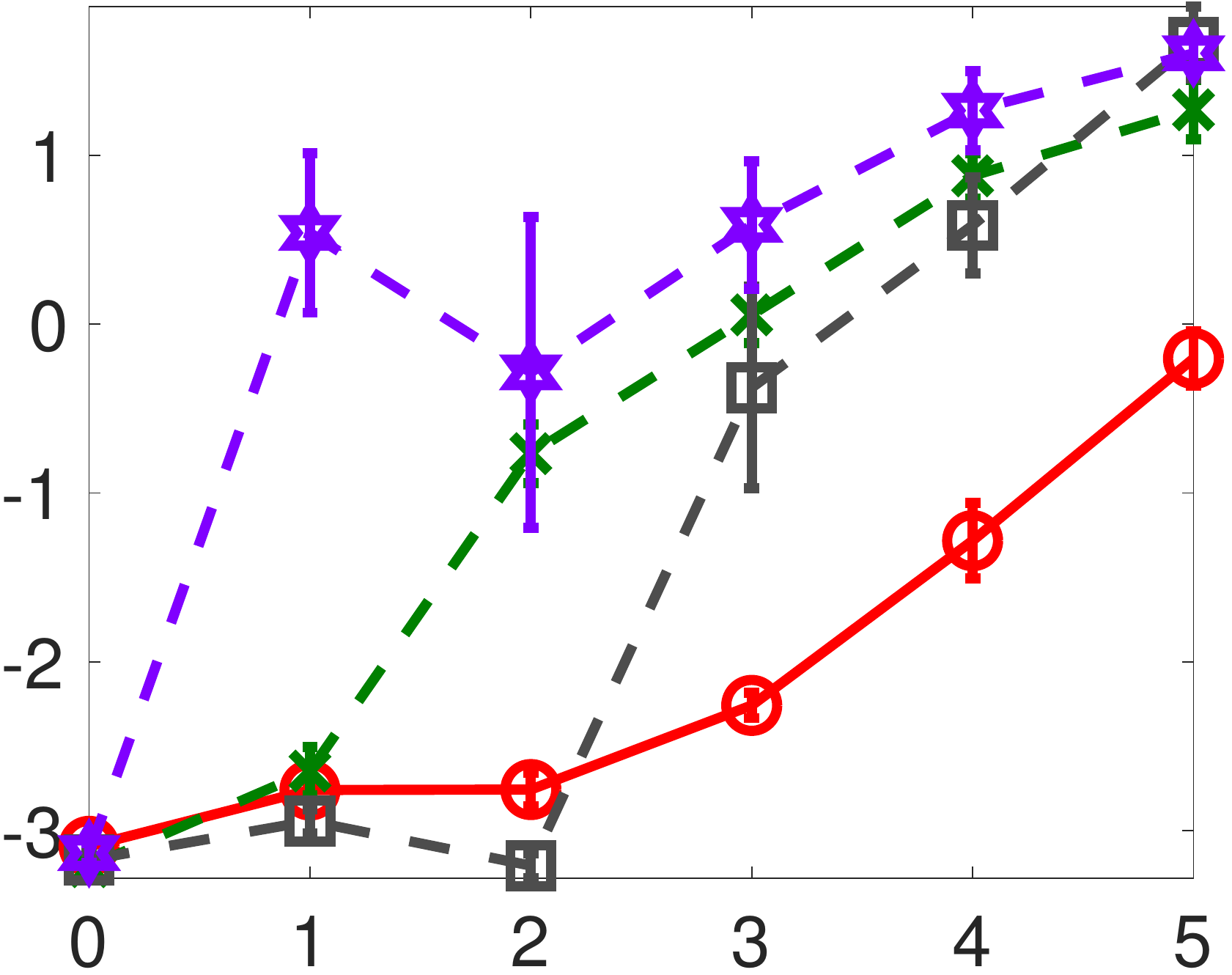} &
\raisebox{1.0em}{\includegraphics[width=0.15\textwidth]{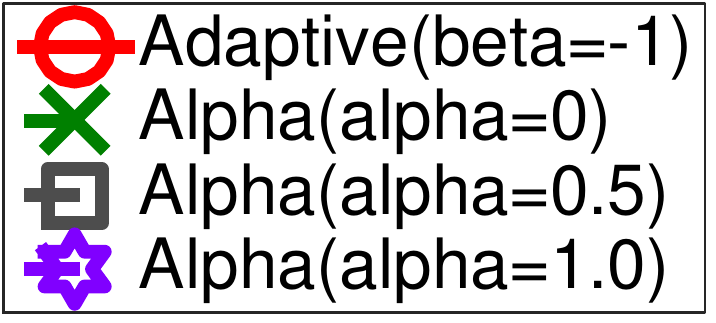}} \\
Non-Gaussianity $s$ & Non-Gaussianity $s$ & Non-Gaussianity $s$  \\
\end{tabular}
\caption{\small Results on randomly generated Gaussian mixture models.
(a) plots the average mode-shift distance;
(b-c) show the MSE of mean and variance.
All results are averaged over 10 random trials.
}
\label{fig:gmm_scale}
\end{figure*}

\subsection{Bayesian Neural Network}
We evaluate our approach on Bayesian neural network regression tasks. 
The datasets are collected from the UCI dataset repository\footnote{\url{https://archive.ics.uci.edu/ml/datasets.html}}.
Similarly to \citet{li2016renyi},
we use a single-layer neural network with 50 hidden units 
and ReLU activation, except that we take 100 hidden units for the Protein and Year
dataset which are relatively large.
We use a fully factorized Gaussian approximation to the
true posterior and 
Gaussian prior 
for neural network weights. 
All datasets are randomly partitioned into $90\%$
for training and $10\%$ for testing.
We use Adam optimizer \citep{kingma2014adam} with a constant learning rate of $0.001$.
The gradient is approximated by $n=100$ draws of $x_i \sim q_\theta$ and a minibatch of size 32 from the training data points.  
All results are averaged over 20 random partitions,
except for Protein and Year, on which $5$ trials are repeated.

We summarize the average RMSE and test log-likelihood with 
standard error in Table~\ref{tbl:bayesnn}. 
We compare our algorithm with $\alpha=0$ (KL divergence) and $\alpha=0.5$, 
which are reported as the best for this task in \citet{li2016renyi}.
More comparisons with different choices of $\alpha$ are provided in the appendix. 
We can see from Table~\ref{tbl:bayesnn} that 
our approach takes advantage of an adaptive choice of $f$-divergence and 
achieves the best performance for both test RMSE and test log-likelihood 
in most of the cases. 


\begin{small}
\begin{table}[t]
\setlength{\tabcolsep}{1.2pt}
\centering
\begin{small}
\begin{tabular}{c|ccc|ccc}
\hline
& \multicolumn{3}{c}{Average Test RMSE}  & \multicolumn{3}{c}{Average Test Log-likelihood} \\ 
\hline
dataset & $\beta=-1.0$ & $\alpha=0.0$ & $\alpha=0.5$  & $\beta=-1.0$ & $\alpha=0.0$ & $\alpha=0.5$ \\
\hline
Boston & $\pmb{2.828\pm0.177}$ & $2.956\pm0.171$ & $2.990\pm0.173$ & $\pmb{-2.476\pm0.177}$ & $-2.547\pm0.171$ & $-2.506\pm0.173$ \\ 
Concrete& $\pmb{5.371\pm0.115}$ & $5.592\pm0.124$ & $5.381\pm0.111$ & $\pmb{-3.099\pm0.115}$ & $-3.149\pm0.124$ & $-3.103\pm0.111$ \\
Energy& $\pmb{1.377\pm0.034}$&$1.431\pm0.029$&$1.531\pm0.047$&$\pmb{-1.758\pm0.034}$&$-1.795\pm0.029$&$-1.854\pm0.047$ \\
Kin8nm& $0.085\pm0.001$&$0.088\pm0.001$&$\pmb{0.083\pm0.001}$&$~~~1.055\pm0.001$&$~~~1.012\pm0.001$&$~~~\pmb{1.080\pm0.001}$ \\
Naval & $\pmb{0.001\pm0.000}$&$\pmb{0.001\pm0.000}$&$0.004\pm0.000$&$~~~\pmb{5.468\pm0.000}$&$~~~5.269\pm0.000$&$~~~4.086\pm0.000$ \\
Combined& $\pmb{4.116\pm0.032}$&$4.161\pm0.034$&$4.154\pm0.042$&$\pmb{-2.835\pm0.032}$&$-2.845\pm0.034$&$-2.843\pm0.042$ \\
Wine& $0.636\pm0.008$&$\pmb{0.634\pm0.007}$&$\pmb{0.634\pm0.008}$&$-0.962\pm0.008$&$\pmb{-0.959\pm0.007}$&$-0.971\pm0.008$\\
Yacht &$\pmb{0.849\pm0.059}$&$0.861\pm0.056$&$1.146\pm0.092$&$\pmb{-1.711\pm0.059}$&$-1.751\pm0.056$&$-1.875\pm0.092$ \\
Protein &$\pmb{4.487\pm0.019}$&$4.565\pm0.026$&$4.564\pm0.040$&$\pmb{-2.921\pm0.019}$&$-2.938\pm0.026$&$-2.928\pm0.040$\\
Year &$\pmb{8.831\pm0.037}$&$8.859\pm0.036$&$8.985\pm0.042$&$-3.570\pm0.037$&$-3.600\pm0.036$&$\pmb{-3.518\pm0.042}$ \\
\hline
\end{tabular}
\end{small}
\caption{\small  Average test RMSE and log-likelihood for Bayesian neural regression.}
\vspace{1em}
\label{tbl:bayesnn}
\end{table}
\end{small}

\subsection{Application in Reinforcement Learning}
\label{sec:rl}
We now demonstrate an application of our method in  reinforcement learning, applying it as an inner loop to improve a recent soft actor-critic(SAC) algorithm \citep{haarnoja2018soft}. 
See more related applications at \citet[e.g.][]{belousov2017f, belousov2019entropic,belousovmean18mean}.
We start with a brief introduction of the background of SAC 
and then test our method in MuJoCo \footnote{\url{http://www.mujoco.org/}} environments.
\paragraph{Reinforcement Learning Background} Reinforcement learning considers the problem of finding optimal policies for agents that interact with uncertain environments to 
maximize the long-term cumulative reward.  
This is formally framed as a Markov decision process 
in which agents iteratively  
take actions $a$ based on observable states $s$,  
and receive a reward signal $r(s,a)$ immediately following the action $a$ performed at state $s$. 
The change of the states is governed by  an unknown environmental dynamic defined by a transition probability $T(s'|s,a)$. 
The agent's action $a$ is selected by a conditional probability distribution $\pi(a|s)$ called policy. 
In policy gradient methods, 
we consider a set of candidate policies $\pi_\theta(a|s)$ parameterized by $\theta$ and obtain the optimal policy by maximizing the expected cumulative reward
$$
J(\theta)
= \E_{s\sim d_\pi, a \sim \pi(a|s)} \left[ r(s,a) \right ], $$
where $d_{\pi}(s) = \sum_{t=1}^\infty \gamma^{t-1} \prob(s_t = s)$ is the 
unnormalized  discounted state visitation distribution 
with discount factor $\gamma\in (0,1)$. 

\paragraph{Soft Actor-Critic}
(SAC) is an off-policy optimization algorithm derived based on maximizing the expected reward with an entropy regularization. It iteratively updates a Q-function $Q(a,s)$, which predicts that cumulative reward of taking action $a$ under state $s$, 
as well as a policy $\pi(a | s)$ which selects action $a$ to maximize the expected value of $Q(s,a)$.  
The update rule of $Q(s,a)$ is based on a variant of Q-learning that matches the Bellman equation, whose detail can be found in \citet{haarnoja2018soft}. 
At each iteration of SAC, the update of policy $\pi$ is achieved by minimizing KL divergence
\begin{align}\label{pinew}
\pi^{\mathrm{new}} = \argmin_{\pi}\E_{s\sim  d}\left [ \KL(\pi(\cdot |s) ~||~ p_Q(\cdot |s)) \right ],  \\
p_Q(a|s) =  \exp \left (\frac{1}{\tau}(Q (a, s)-V(s))\right), 
\end{align}
where 
$\tau$ is a temperature parameter, and $V(s) = \tau\log \int_{a} \exp(Q(a,s)/\tau)$, serving as a normalization constant here, is a soft-version of value function and is also iteratively updated in SAC. Here, $d(s)$ is a visitation distribution on states $s$, which is taken to be the empirical distribution of the states in the current replay buffer in SAC. 
We can see that \eqref{pinew} can be viewed as a variational inference problem on conditional distribution $p_Q(a|s)$, with the typical KL objective function ($\alpha = 0$).

\begin{figure*}[t]
\centering
{
\renewcommand{\arraystretch}{1} 
\begin{tabu}{ccccc}
\small{Ant} &  \small{HalfCheetah} & \small{Humanoid(rllab)} \\
\raisebox{1.0em}{\rotatebox{90}{\small Average Reward}}
\includegraphics[width=.25\textwidth]{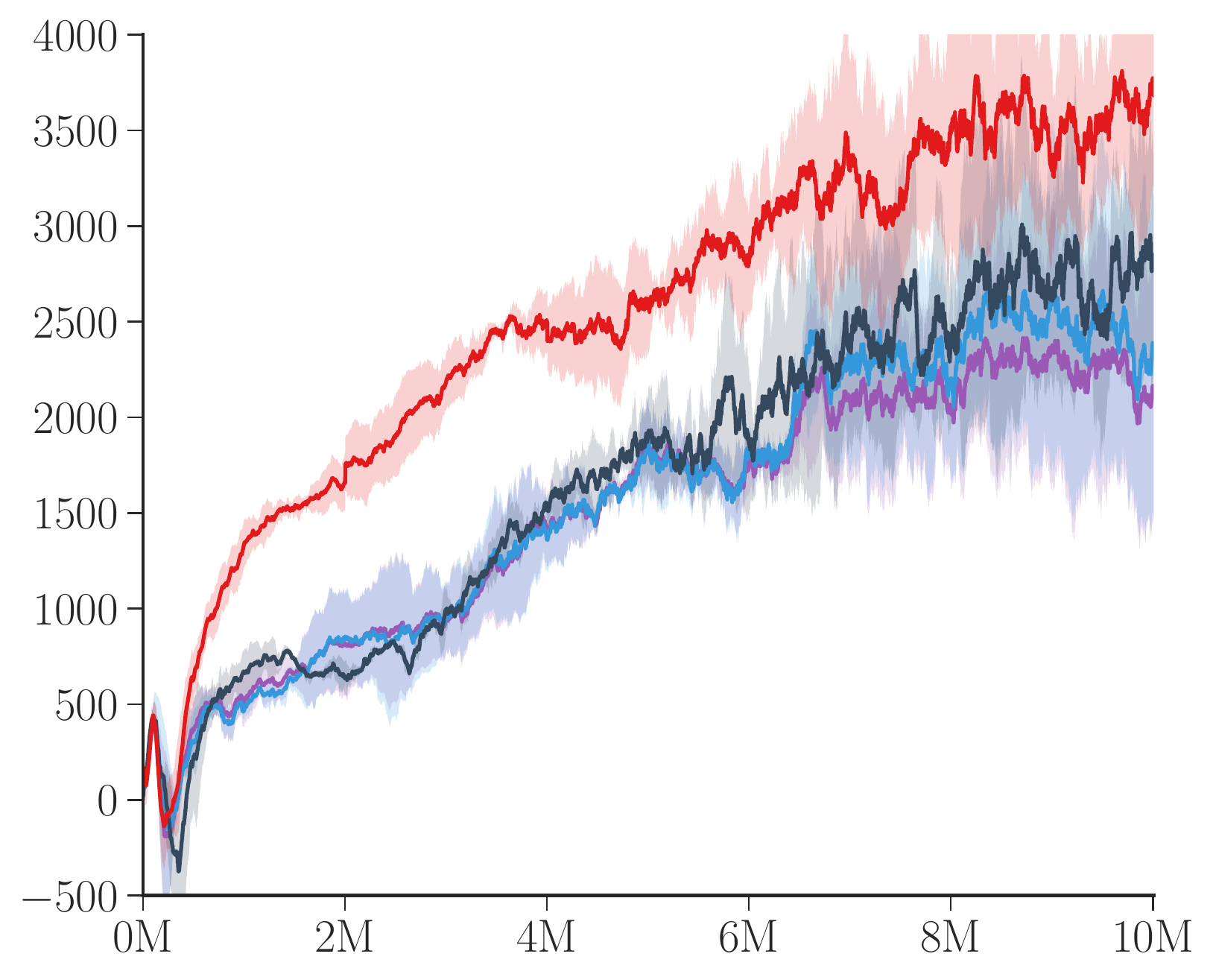}&
\includegraphics[width=.25\textwidth]{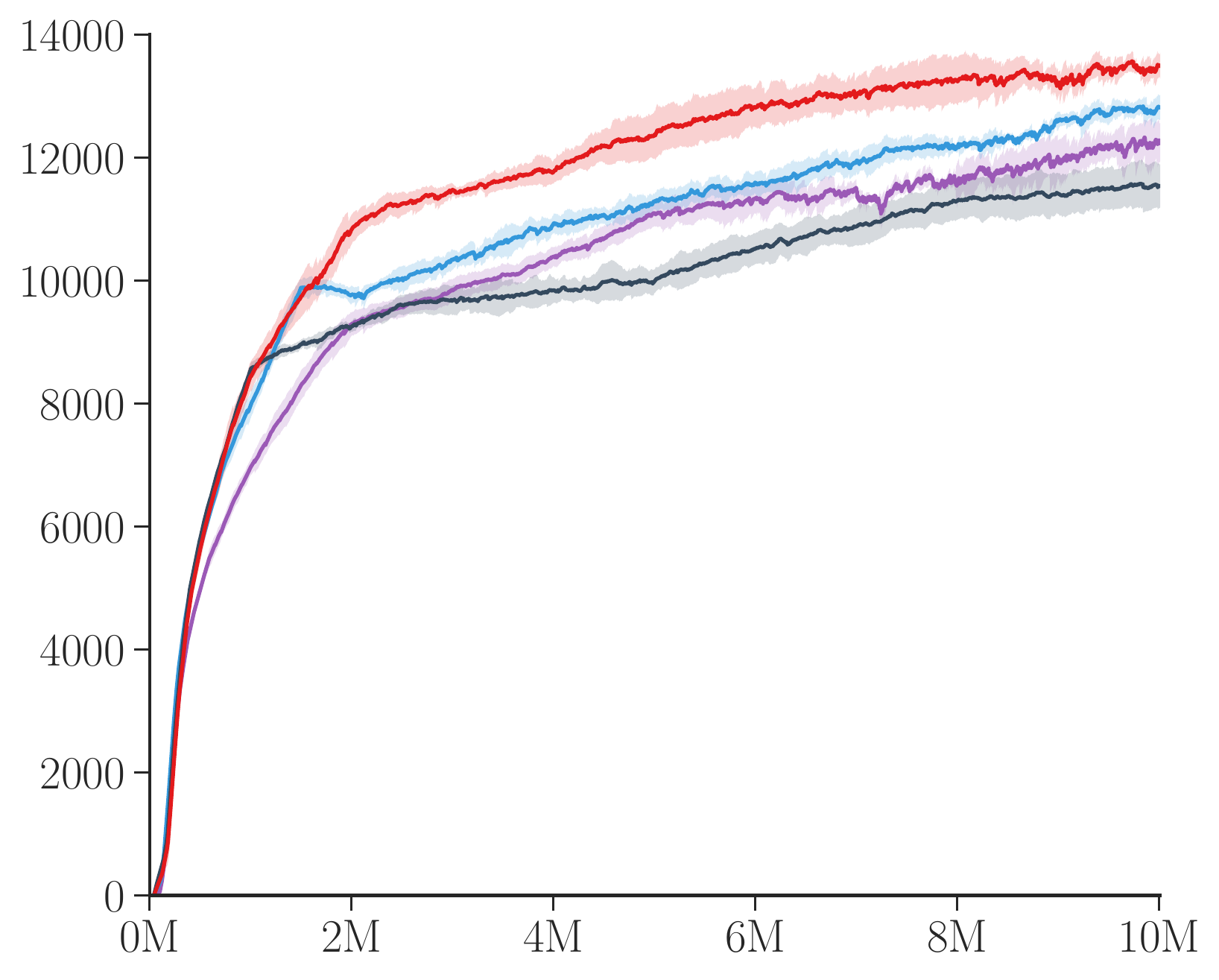}&
\includegraphics[width=.25\textwidth]{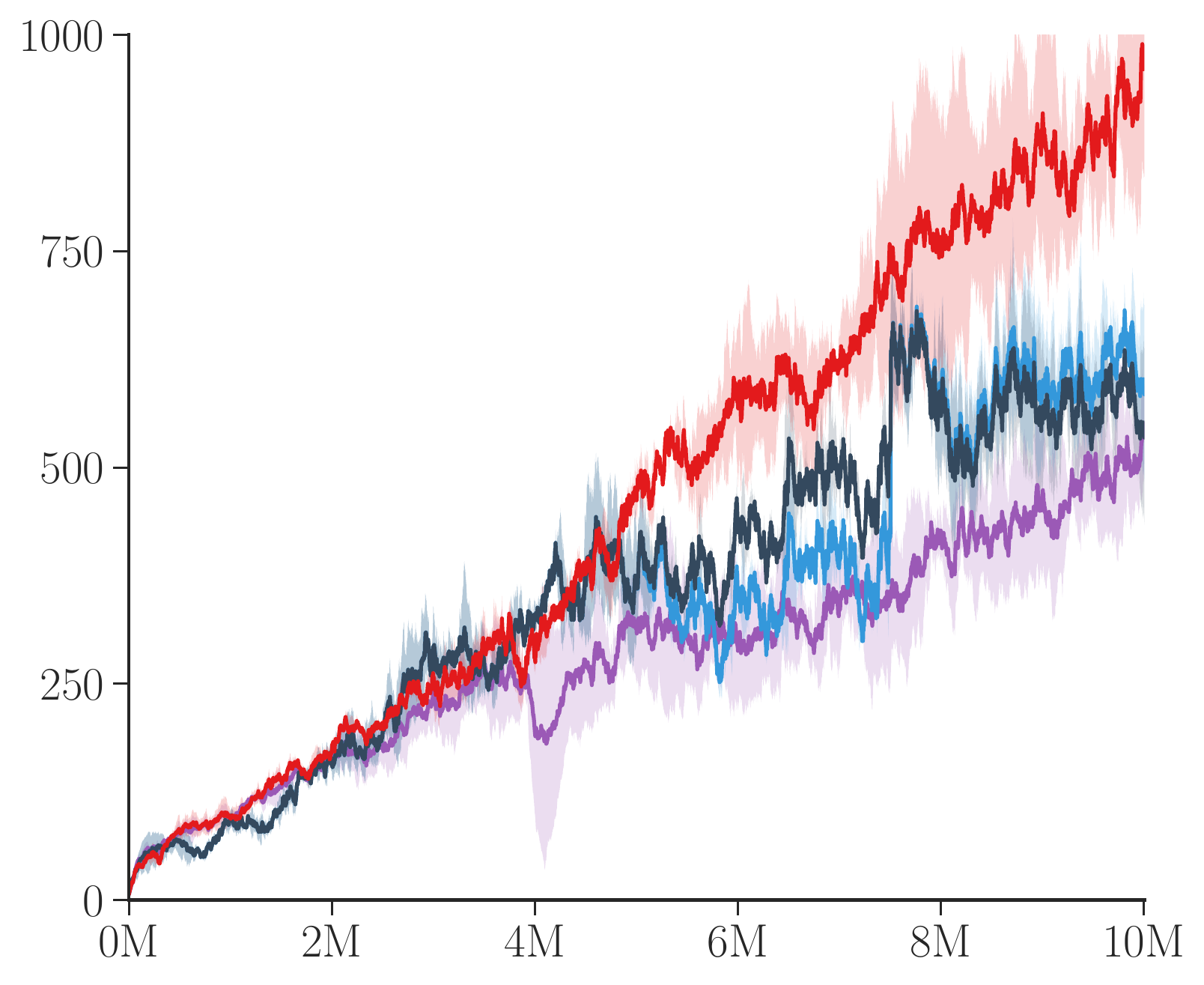}\\
\small{Walker} &  \small{Hopper} & \small{Swimmer(rllab)} \\
\raisebox{1.0em}{\rotatebox{90}{\small Average Reward}}
\includegraphics[width=.25\textwidth]{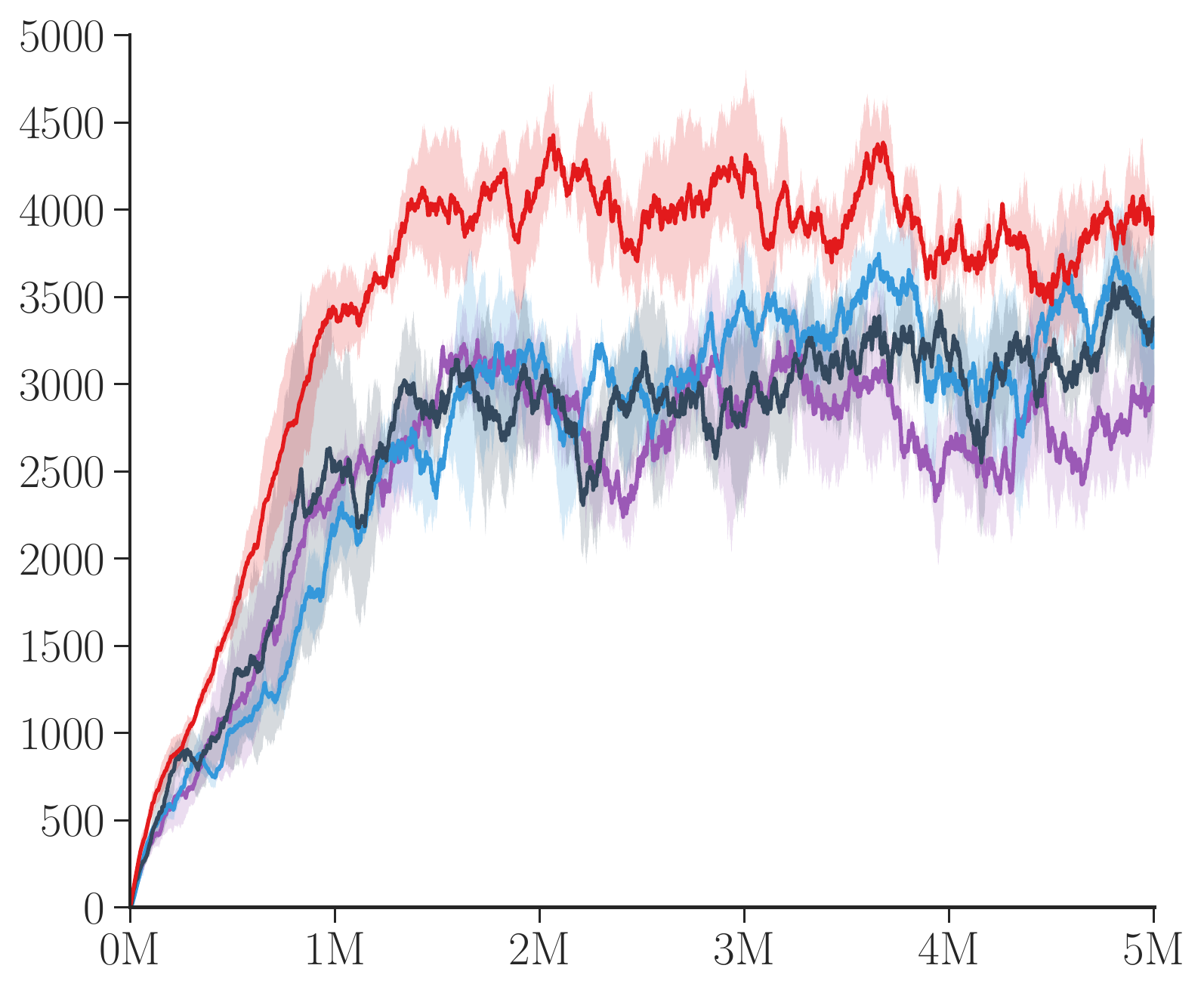} &
\includegraphics[width=.25\textwidth]{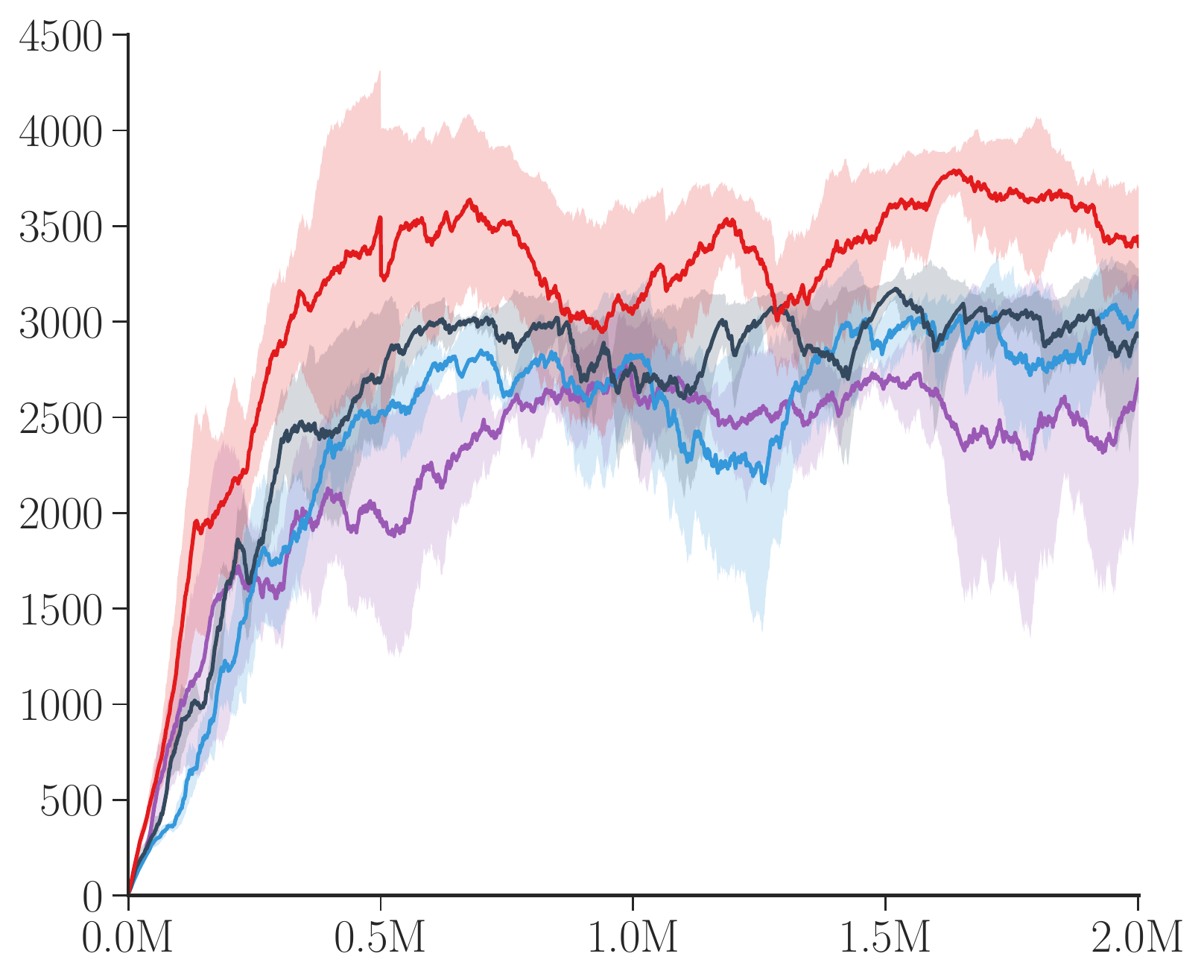} &
\includegraphics[width=.25\textwidth]{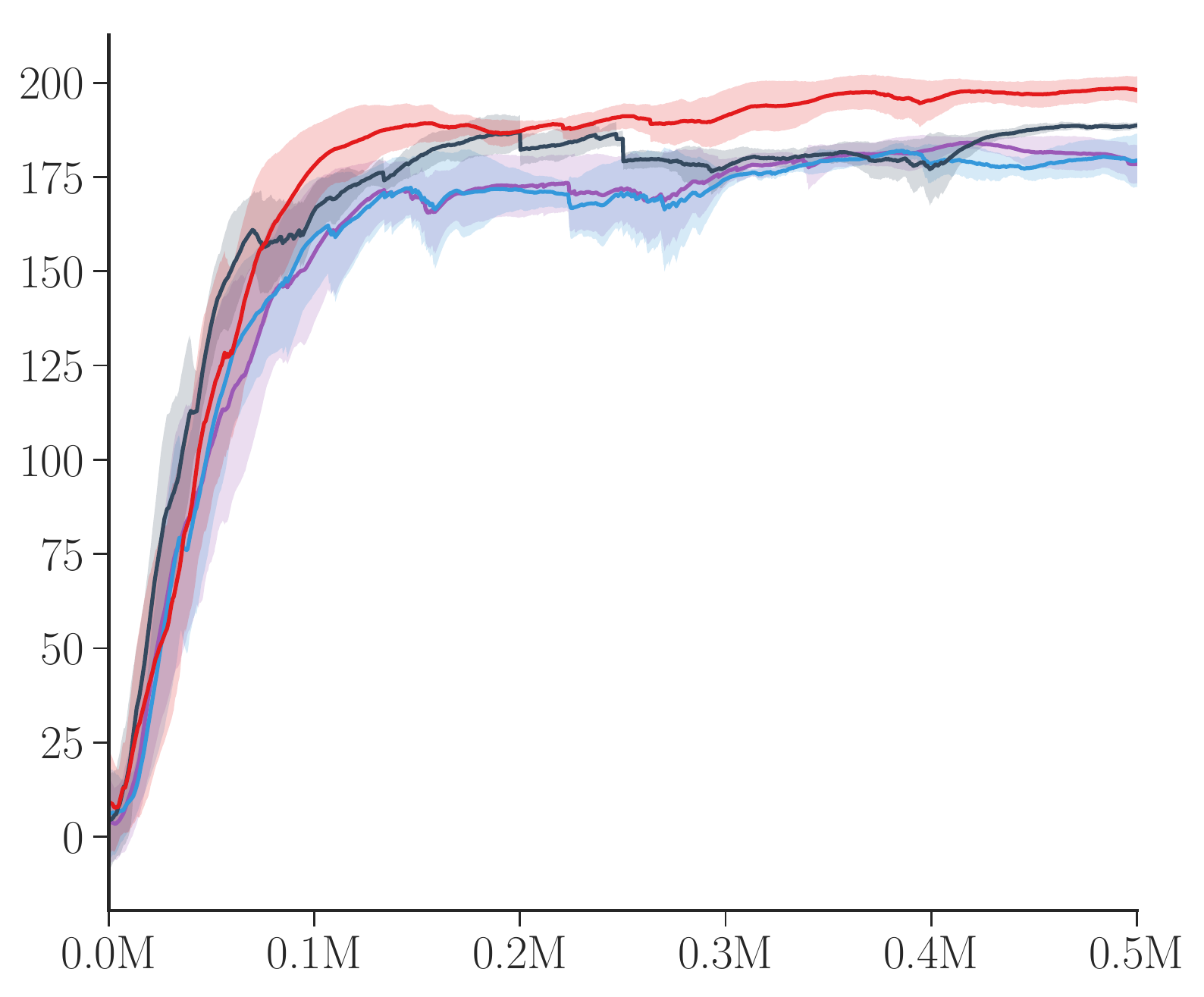}
\llap{\raisebox{1.2em}{
\includegraphics[height=1.2cm]{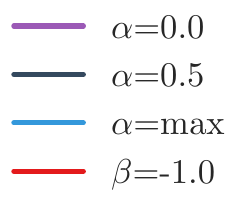}}} \\
\end{tabu}}
\vspace{-10bp}
\caption{\small Soft Actor Critic (SAC) with policy updated by Algorithm~\ref{alg1} with $\beta=-1$, or $\alpha$-divergence VI with different $\alpha$ ($\alpha=0$ corresponds to the original SAC). The reparameterization gradient estimator is used in all the cases.
In the legend, ``$\alpha=\max$'' denotes setting $\alpha=+\infty$ in $\alpha$-divergence. 
}
\label{fig:different_f}
\end{figure*}

\paragraph{SAC With Tail-adaptive $f$-Divergence}
To apply $f$-divergence, we first rewrite \eqref{pinew} to transform the conditional distributions to joint distributions.  
We define joint distribution $p_Q(a,s) = \exp((Q(a,s)-V(s))/\tau) d(s)$ and $q_\pi(a,s) = \pi(a|s)d(s)$, then we can show that $\E_{s\sim d}[\KL(\pi(\cdot|s)~||~p_Q(\cdot|s))] = \KL(q_\pi ~||~p_Q)$. 
This motivates us to extend the objective function in \eqref{pinew} to more general $f$-divergences,  
$$
D_f(p_Q ~||~ q_\pi ) = \E_{s\sim d} \E_{a|s\sim \pi} 
\left [ f \bigg(\frac{\exp((Q(a,s)-V(s))/\tau)}{\pi(a|s)}\bigg)  -  f(1) \right ]. 
$$
By using our tail-adaptive $f$-divergence, we can readily 
apply our Algorithm~\ref{alg1} (or Algorithm~2 in the Appendix) to update $\pi$ in SAC, allowing us to obtain $\pi$ that counts the multi-modality of $Q(a,s)$ more efficiently. Note that the standard $\alpha$-divergence with a fixed $\alpha$ also yields a new variant of SAC that is not yet studied in the literature.  

\paragraph{Empirical Results} 
We follow the  experimental setup of \citet{haarnoja2018soft}.
The policy $\pi$, the value function $V(s)$, and 
the Q-function $Q(s,a)$ are neural networks
with two fully-connected layers of 128 hidden units each.
We use Adam  \citep{kingma2014adam} with a constant learning rate of $0.0003$ for optimization. 
The size of the replay buffer for HalfCheetah is $10^7$, and
we fix the size to $10^6$ on other environments in a way similar to \citet{haarnoja2018soft}.

We compare with the original SAC ($\alpha =0$), and
also other $\alpha$-divergences, such as $\alpha=0.5$ and $\alpha=\infty$ (the $\alpha = \max$ suggested in \citet{li2016renyi}).
Figure~\ref{fig:different_f} summarizes the total average reward
of evaluation rollouts during training on various MuJoCo environments. 
For non-negative $\alpha$ settings, methods with larger $\alpha$ give higher
average reward than the original KL-based SAC in most of the cases.
Overall, our adaptive $f$-divergence
substantially outperforms all other $\alpha$-divergences
on all of the benchmark tasks in terms of the final performance, 
and learns faster than all the baselines in most environments. 
We find that our improvement is especially significant on high dimensional and complex environments like Ant and Humanoid. 

%% file: tex/appendix.tex
\clearpage
\appendix

\section{Proof of Proposition~4.1}
\begin{proof}
Taking $h(t) = f''(t)$, $a = f'(0)$ and $b = f(0)$ in Eq~\eqref{fff}, we have 
\begin{align*}
& (f'(0)t+ f(0)) ~+~ \int_0^\infty (t-\mu)_+ h(\mu)d\mu   \\
& = (f'(0)t+ f(0)) ~+~  \int_0^t (t-\mu) f''(\mu) d\mu \\
& = (f'(0)t+ f(0)) ~+~  (t-\mu) f'(\mu) \bigg |_{\mu=0}^t + \int_0^t f'(\mu) d \mu  \text{~~~~~~\small (integration by parts)}\\ 
& =  f(0) + \int_0^t f'(\mu) d \mu \\
& = f(t). 
\end{align*}
Conversely, if $f(t) = (at+b) + \int_0^\infty (t-\mu)_+ h(\mu)d\mu$, calculation shows 
$$
f'(t) = a + \int_0^t h(\mu) d\mu, ~~~f''(t) = h(t).
$$
Therefore, $f$ is convex if $h$ is non-negative. 

To prove Eq.~\eqref{eq:f_sec_dev}, 
we substitute $f(t) = f'(0)t+ f(0) ~+~ \int_0^\infty (t-\mu)_+ f''(\mu)d\mu $ into 
the definition of $f$-divergence, 
\begin{align*}
D_f(p~||~q)
& = \E_{q} \left [f\left (\frac{p(x)}{q(x)}\right ) - f(1)\right]  \\
& = \E_{q} \left [f'(0)\frac{ p(x)}{q(x)}+f(0) ~+~ \int_0^\infty  (p(x)/q(x) - \mu)_+ f''(\mu) d\mu  - f(1)\right ] \\
& = [f'(0) + f(0) - f(1) ]~+~ \int_0^\infty \E_q\left [ \left ( \frac{p(x)}{q(x)} - \mu\right )_+\right ] f''(\mu) d\mu. 
\end{align*}
This completes the proof. 
\end{proof}

\section{Proof of Proposition~\ref{pro:gradient}}
\begin{proof}
By chain rule and the ``score-function trick'' $\nabla_\theta q_\theta(x) = q_\theta(x) \nabla_\theta \log q_\theta(x)$, we have 
\begin{align*}
\nabla_\theta D_f(p ~||~ q_\theta) 
&= \E_{q_\theta}\left [ \nabla_\theta f\bigg( \frac{p(x)}{q_\theta (x)}\bigg) + f\bigg( \frac{p(x)}{q_\theta (x)}\bigg) \nabla_\theta \log q_\theta(x)\right ] \\
&= \E_{q_\theta}\left [  f'\bigg( \frac{p(x)}{q_\theta (x)}\bigg) \nabla_\theta
\left (\frac{p(x)}{q_\theta(x)}\right ) + f\bigg( \frac{p(x)}{q_\theta (x)}\bigg) \nabla_\theta \log q_\theta(x)\right ] \\
&= \E_{q_\theta}\left [ - f'\bigg( \frac{p(x)}{q_\theta (x)}\bigg)  \bigg (\frac{p(x)}{q_\theta(x)}\bigg ) \nabla_\theta \log q_\theta(x) + f\bigg( \frac{p(x)}{q_\theta (x)}\bigg) \nabla_\theta \log q_\theta(x)\right ] \\
& = - \E_{q_\theta} \left [ \rho_f\left ( \frac{p(x)}{q_\theta(x)}\right )\log q_\theta(x) \right],
\end{align*}
where $\rho_f(t) = f'(t) t - f(t)$. 
This proves Eq.~\eqref{score}.  

To prove Eq.~\eqref{eq:repa}, we note that for any function $\phi$, we have 
by the \emph{reparamertization trick}:  
\begin{align*}
\nabla_\theta\E_{q_\theta}[ \phi(x) ] 
&= \E_{x\sim q_\theta} [\phi(x) \nabla_\theta \log q_\theta(x)] ~~~~~\text{(score function)}\\
&= \E_{\xi \sim q_0} [\nabla_x \phi(x) \nabla_\theta g_\theta(\xi)] ~~~\text{(reparameterization trick)}, 
\end{align*}
where we assume $x
\sim q_\theta$ is generated by $x = g_\theta(\xi)$, ~ $\xi\sim q_0$. 

Taking $\phi(x)  = \rho_f(p(x)/q_\theta(x))$ in Eq.~\eqref{score}, we have 
$$
\begin{aligned}
\nabla_\theta D_f(p~||~q_\theta) &= -\E_{x\sim q_\theta}\bigg[ \rho_f\bigg( \frac{p(x)}{q_\theta(x)}\bigg) \nabla_\theta \log q_\theta(x) \bigg] \\
&= -\E_{\xi \sim q_0} \bigg[\nabla_x \rho_f\bigg( \frac{p(x)}{q_\theta(x)}\bigg) \nabla_\theta g_\theta(\xi) \bigg] \\
&= -\E_{\xi \sim q_0} \bigg[  \rho'_f\bigg( \frac{p(x)}{q_\theta(x)}\bigg) \nabla_x \bigg(\frac{p(x)}{q_\theta(x)} \bigg) \nabla_\theta g_\theta(\xi) \bigg]\\
&= -\E_{\xi \sim q_0} \bigg[  \rho'_f\bigg( \frac{p(x)}{q_\theta(x)}\bigg) \bigg( \frac{p(x)}{q_\theta(x)} \bigg)\nabla_x \log\bigg(\frac{p(x)}{q_\theta(x)} \bigg) \nabla_\theta g_\theta(\xi) \bigg]\\
&= -\E_{\xi \sim q_0} \bigg[  \omeganew_f\bigg( \frac{p(x)}{q_\theta(x)} \bigg)
\nabla_x \log\bigg(\frac{p(x)}{q_\theta(x)}\bigg) \nabla_\theta g_\theta(\xi) \bigg ], 
\end{aligned}
$$
where $\omeganew_f(t) = \rho'_f(t)t$.

\end{proof}

\section{Tail-adaptive $f$-divergence with Score-Function Gradient Estimator}
Algorithm~\ref{alg2} summarizes our method using the score-function gradient estimator \eqref{score}.

\begin{algorithm}[ht] 
\caption{Variational Inference with Tail-adaptive $f$-Divergence (with Score Function Gradient)}
\label{alg2} 
\begin{algorithmic} 
    \STATE Goal: Find the best approximation of $p(x)$ from $\{q_\theta \colon \theta\in \Theta\}$. 
    \STATE Initialize $\theta$, set an index $\beta$ (e.g., $\beta = -1$).
    \FOR{iteration}
    \STATE Draw $\{x_i\}_{i=1}^n \sim q_\theta$. Set  
    $\hat{\bar F}(t) = \sum_{j=1}^n\ind(p(x_j)/q(x_j) \geq t)/n$, and 
     $\rho_i = \hat{\bar F}(p(x_i)/q(x_i))^{\beta}.$
    \STATE Update $\theta \gets \theta  + \epsilon  \Delta \theta$, where $\epsilon$ is stepsize, and 
 \begin{align*}
     \Delta \theta = \frac{1}{z_\rho}\sum_{i=1}^n\left [ \rho_i \nabla_\theta \log q_\theta(x_i) \right ],
 \end{align*}   
where $z_\rho = \sum_{i=1}^n \rho_i$.      
   \ENDFOR
\end{algorithmic}
\end{algorithm}


\newpage \clearpage
\section{More Results for Bayesian Neural Network} 
Table~\ref{tbl:more_bnn_results} shows more results in Bayesian networks with more choices of $\alpha$ in $\alpha$-divergence. 
We can see that our approach achieves 
the best performance in most of the cases.

\begin{table}[ht] 
\setlength{\tabcolsep}{3pt}
\centering
\begin{tabular}{l|cc|cccccc}
\hline
& \multicolumn{8}{c}{Average Test RMSE} \\ 
\hline
Dataset & $\beta=-1.0$ & $\beta=-0.5$ &  $\alpha=-1$ & $\alpha=0$ & $\alpha=0.5$ & $\alpha=1.0$ & $\alpha=2.0$  & $\alpha=+\infty$ \\
\hline
Boston & \textbf{2.828}& 2.948&  3.026&    2.956& 2.990& 2.937& 2.981& 2.985\\
Concrete & \textbf{5.371}& 5.505&   5.717&     5.592& 5.381& 5.462& 5.499& 5.481\\
Energy & \textbf{1.377}& 1.461&      1.646&    1.431& 1.531& 1.413& 1.458& 1.458\\
Kin8nm & 0.085& 0.088&    0.087&     0.088& \textbf{0.083}& 0.084& 0.084& \textbf{0.083}\\
Naval & \textbf{0.001}& \textbf{0.001}&     0.003&    \textbf{0.001}& 0.004& 0.005& 0.004& 0.004\\
Combined & \textbf{4.116}& 4.146&   4.156&     4.161& 4.154& 4.135& 4.188& 4.145\\
Wine & 0.636& \textbf{0.632}&  \textbf{0.632}&    0.634& 0.634& 0.633& 0.635& 0.634\\
Yacht & 0.849& \textbf{0.788}&   1.478&    0.861& 1.146& 1.221& 1.222& 1.234\\
Protein & \textbf{4.487}& 4.531&   4.550&   4.565& 4.564& 4.658& 4.777& 4.579\\
Year & \textbf{8.831}& 8.839&    8.841&   8.859& 8.985& 9.160& 9.028& 9.086\\
\hline\hline
& \multicolumn{8}{c}{Average Test Log-likelihood} \\ 
\hline
dataset & $\beta=-1.0$ & $\beta=-0.5$ & $\alpha=-1$ & $\alpha=0$ & $\alpha=0.5$ & $\alpha=1.0$ & $\alpha=2.0$ & $\alpha=+\infty$ \\
\hline
Boston & \textbf{-2.476}& -2.523&    -2.561&     -2.547& -2.506& -2.493& -2.516& -2.509\\
Concrete & \textbf{-3.099}& -3.133&  -3.171&      -3.149& -3.103& -3.106& -3.116& -3.109\\
Energy & \textbf{-1.758}& -1.814&  -1.946&     -1.795& -1.854& -1.801& -1.828& -1.832\\
Kin8nm & ~1.055& ~1.017&   ~1.024&      ~1.012& ~1.080& ~1.075& ~1.074& ~\textbf{1.085}\\
Naval & ~\textbf{5.468}& ~5.347&  4.178&     ~5.269& ~4.086& ~4.022& ~4.077& ~4.037\\
Combined & \textbf{-2.835}& -2.842&  -2.845&    -2.845& -2.843& -2.839& -2.850& -2.842\\
Wine & -0.962& \textbf{-0.956}&   -0.961&   -0.959& -0.971& -0.968& -0.972& -0.971\\
Yacht & \textbf{-1.711}& -1.718&    -2.201&   -1.751& -1.875& -1.946& -1.963& -1.986\\
Protein & \textbf{-2.921}& -2.930&   -2.934&   -2.938& -2.928& -2.930& -2.947& -2.932\\
Year & -3.570& -3.597&  -3.599&    -3.600& \textbf{-3.518}& -3.529& -3.524& -3.524\\
\hline
\end{tabular}
\caption{Test RMSE and LL results for Bayesian neural network regression.}
\label{tbl:more_bnn_results}
\end{table}

\section{Reinforcement Learning}
In this section, we provide more information and results of the Reinforcement learning experiments, including comparisons of algorithms using score-function gradient estimators (Algorithm~\ref{alg2}).   

\subsection{MuJoCo Environments}
\label{sec:mujoco-envs}
\begin{figure*}[ht]
\centering
{
\setlength{\tabcolsep}{1pt} 
\renewcommand{\arraystretch}{1} 
\begin{tabu}{cccccc}
\includegraphics[width=.15\textwidth]{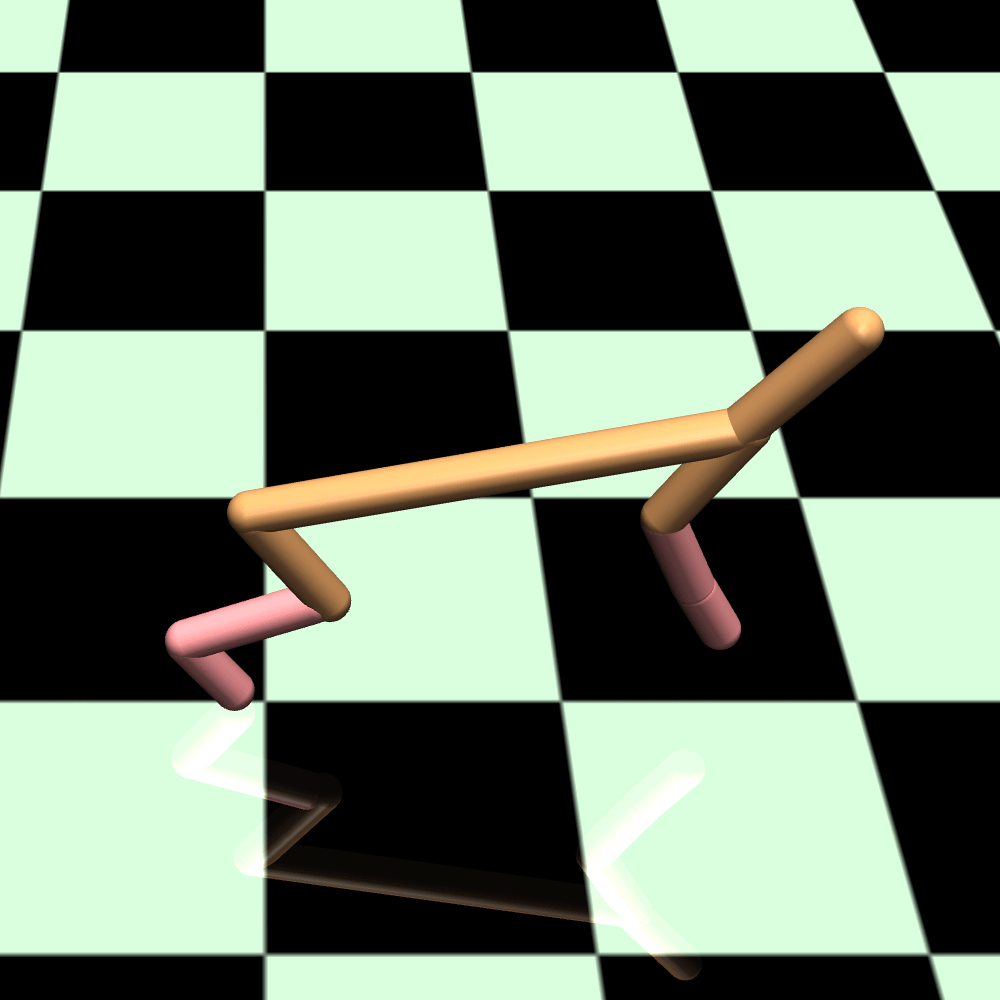}&
\includegraphics[width=.15\textwidth]{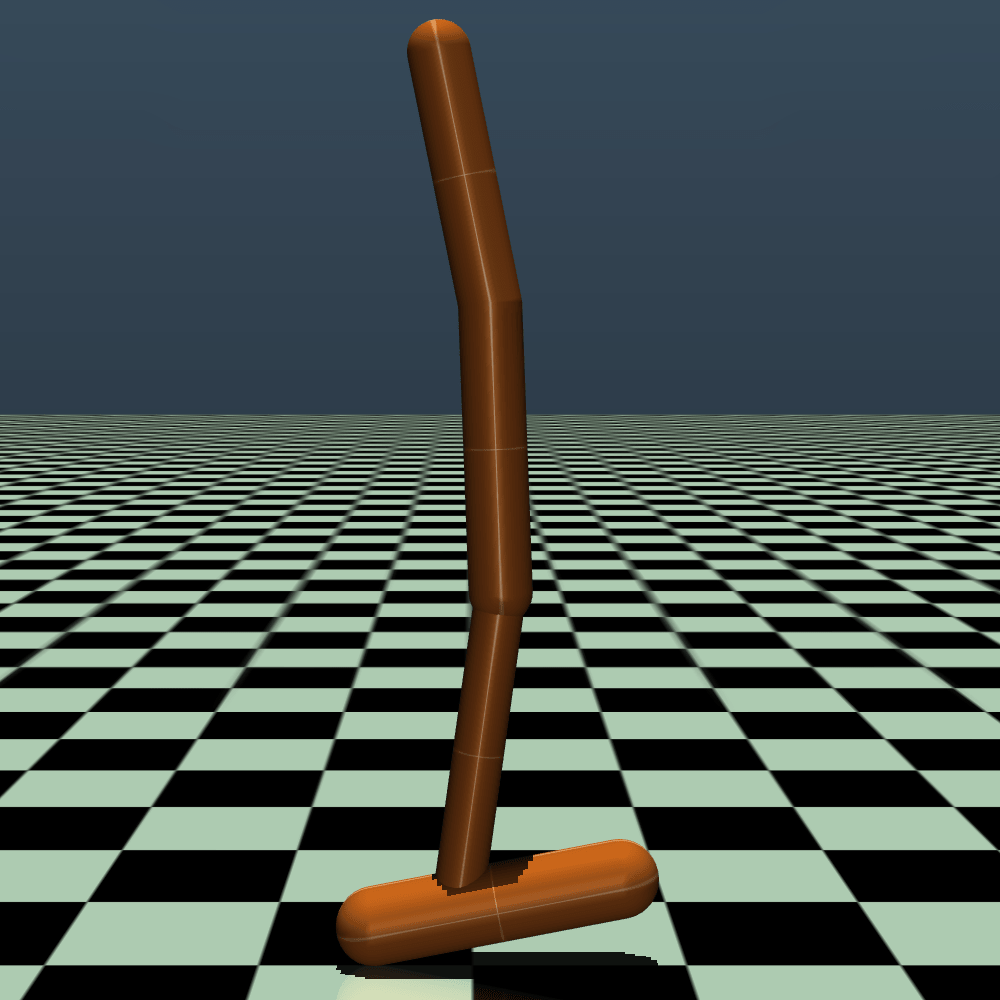}&
\includegraphics[width=.15\textwidth]{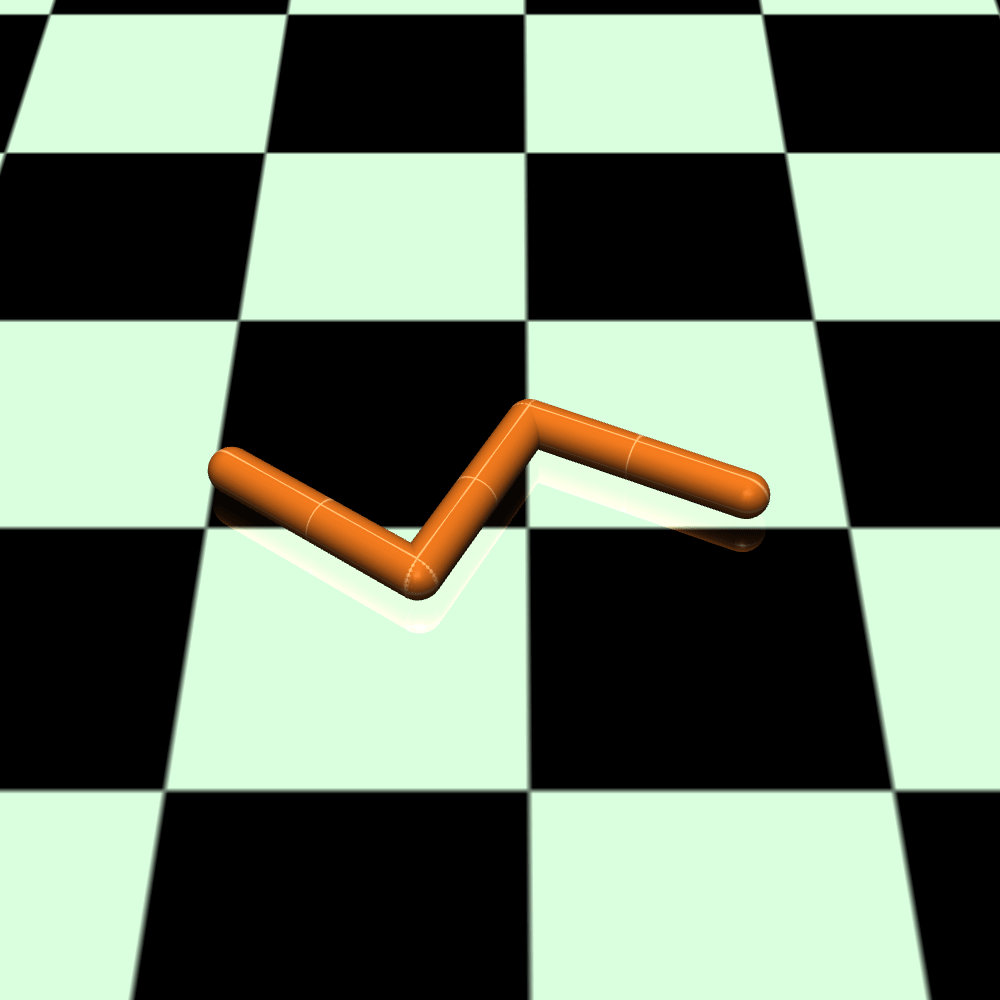}&
\includegraphics[width=.15\textwidth]{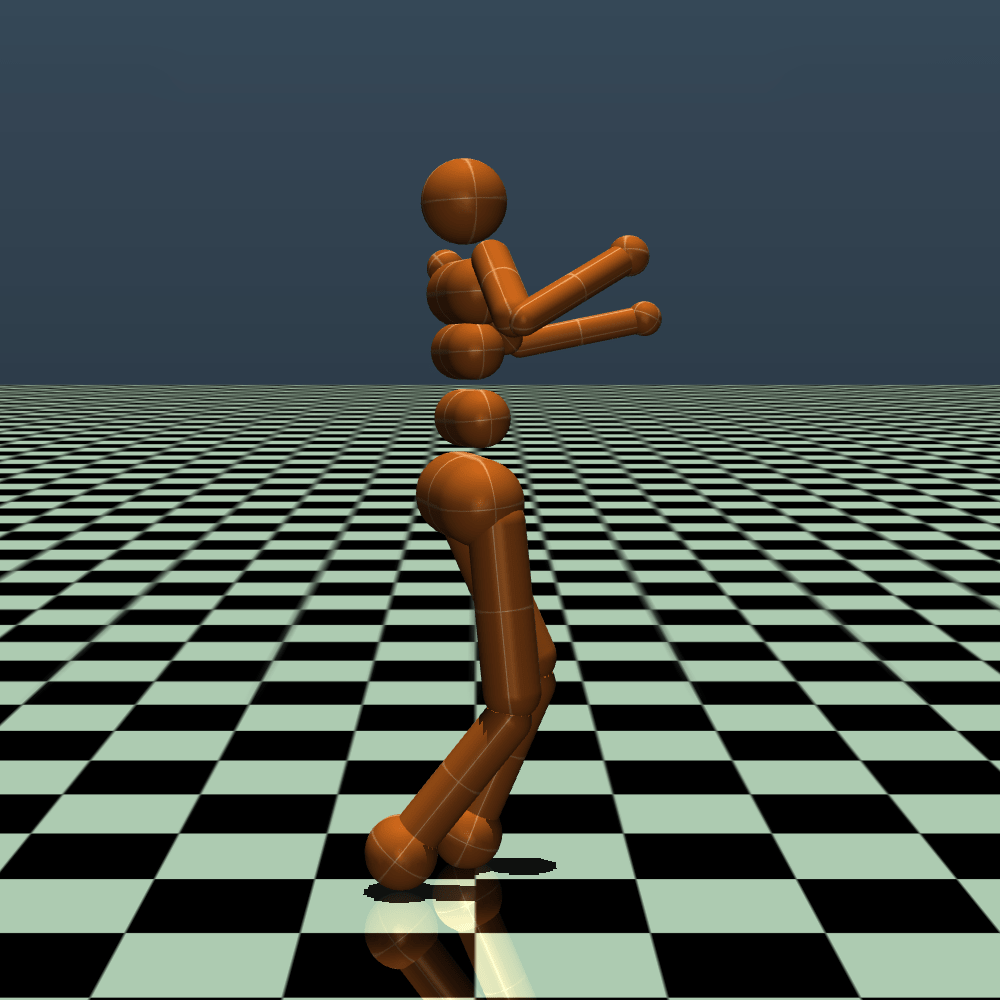}&
\includegraphics[width=.15\textwidth]{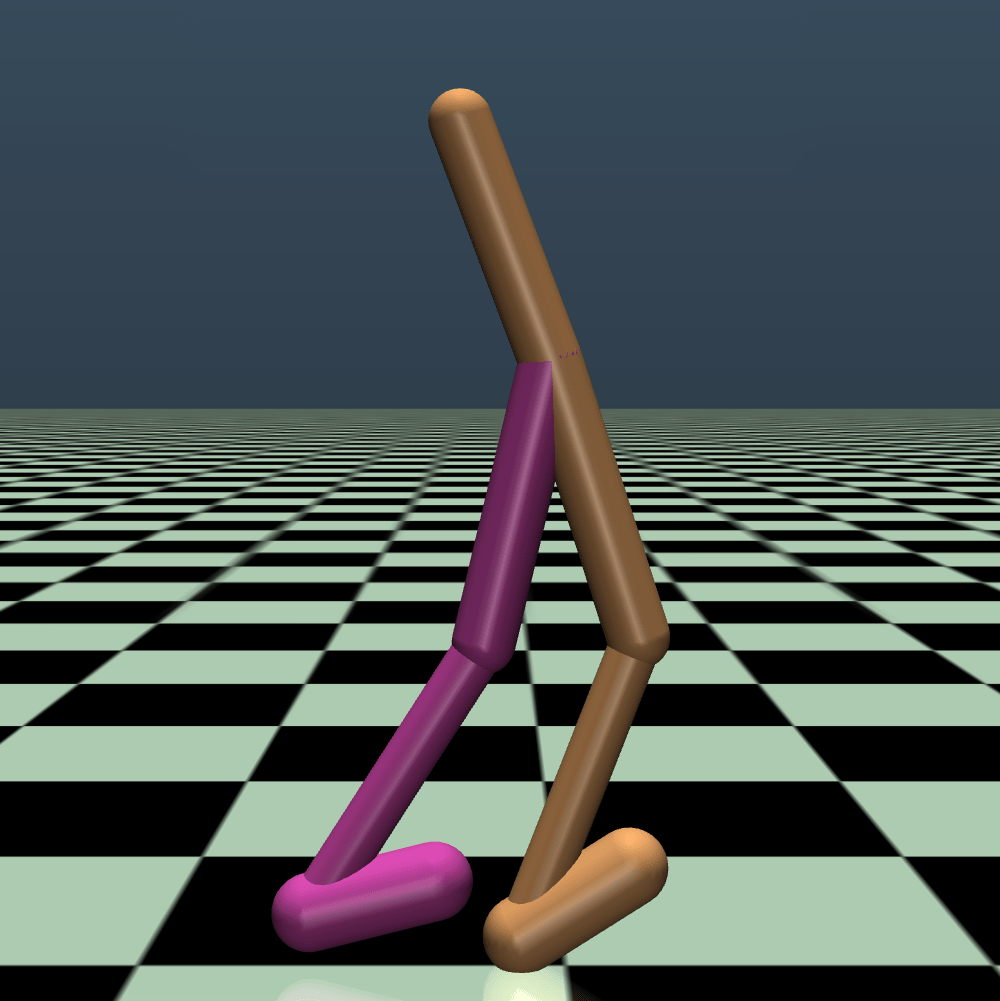}&
\includegraphics[width=.15\textwidth]{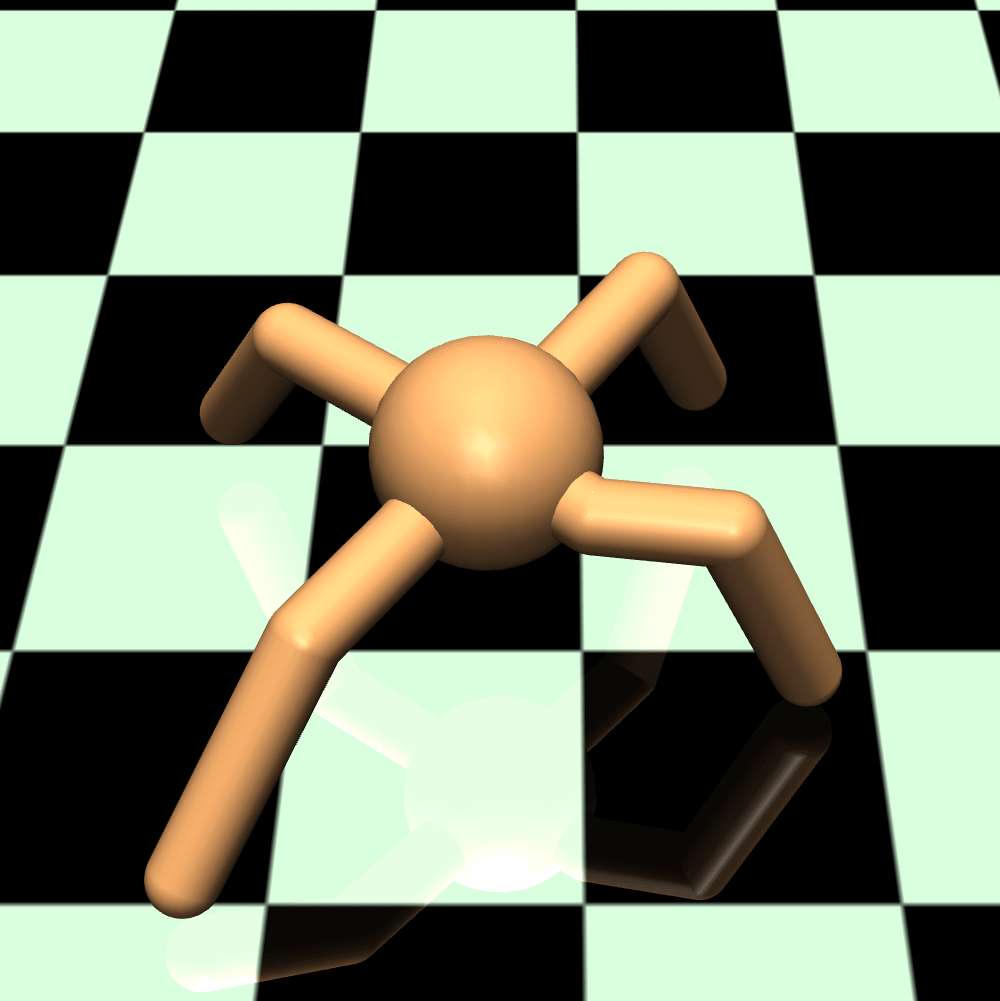}\\
\end{tabu}}
\vspace{-5bp}
\caption{\small{MuJoCo environments used in our reinforcement learning experiments. From left to right: HalfCheetah, Hopper, Swimmer(rllab), Humanoid(rllab), Walker, and Ant.}}
\label{fig:demo}
\end{figure*}

We test six MuJoCo environments in this paper: \emph{HalfCheetah}, \emph{Hopper}, \emph{Swimmer(rllab)}, \emph{Humanoid(rllab)}, \emph{Walker}, and \emph{Ant},
for which the dimensions of the action space are 6, 3, 2, 21, 6, 8, respectively. 
Figure~\ref{fig:demo} shows 
examples of the environment used in our experiments. 

\subsection{Different Choices of $\alpha$}

\begin{figure*}[ht]
\centering
{
\setlength{\tabcolsep}{1pt} 
\renewcommand{\arraystretch}{1} 
\begin{tabu}{ccccc}
\footnotesize{Hopper\tiny{(Score function)}} & 
\footnotesize{Hopper\tiny{(Reparameterization)}} & 
\footnotesize{Walker\tiny{(Score function)}} &
\footnotesize{Walker\tiny{(Reparameterization)}} \\
\raisebox{1.0em}{\rotatebox{90}{\scriptsize Average Reward}}
\includegraphics[width=.23\textwidth]{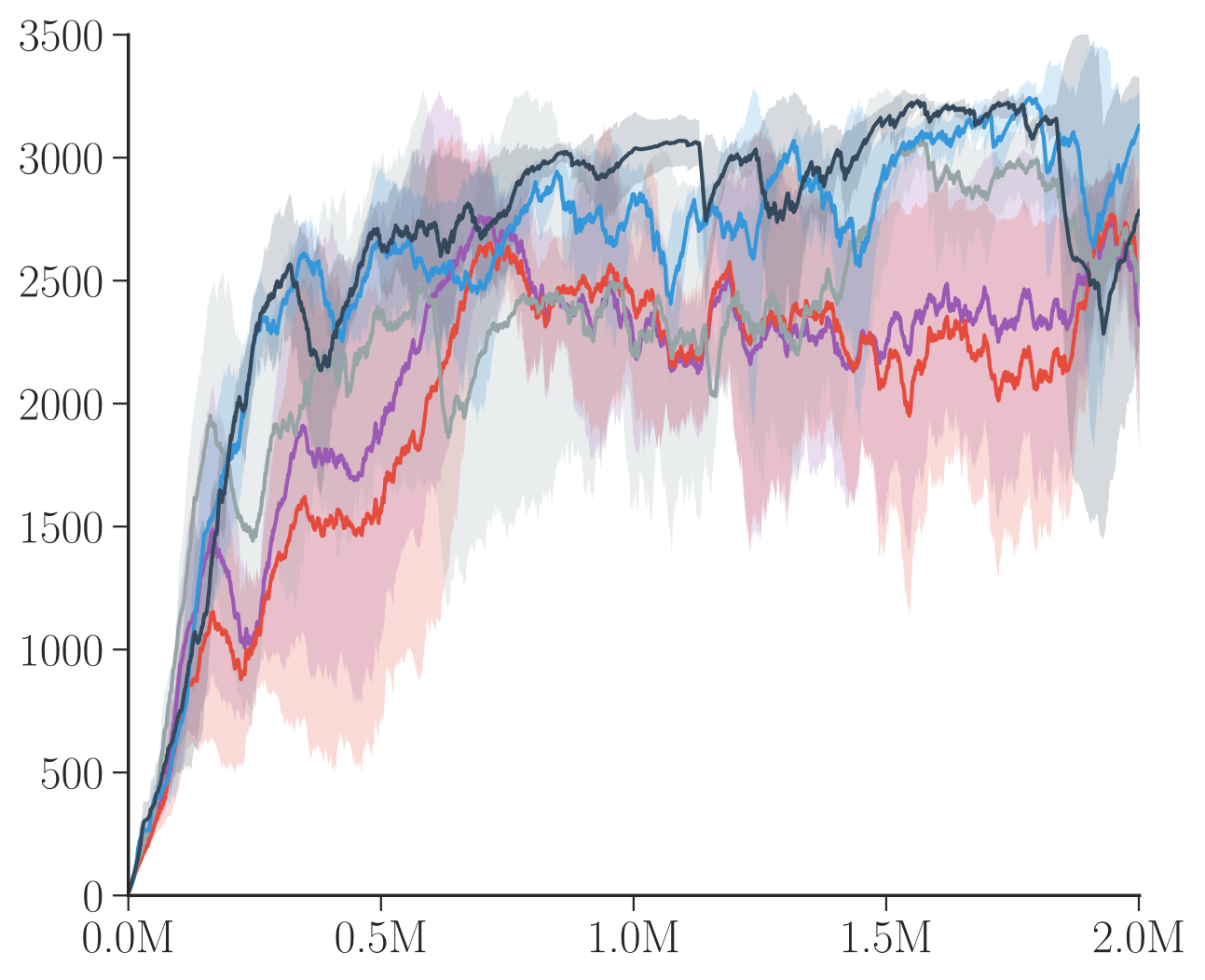} &
\includegraphics[width=.23\textwidth]{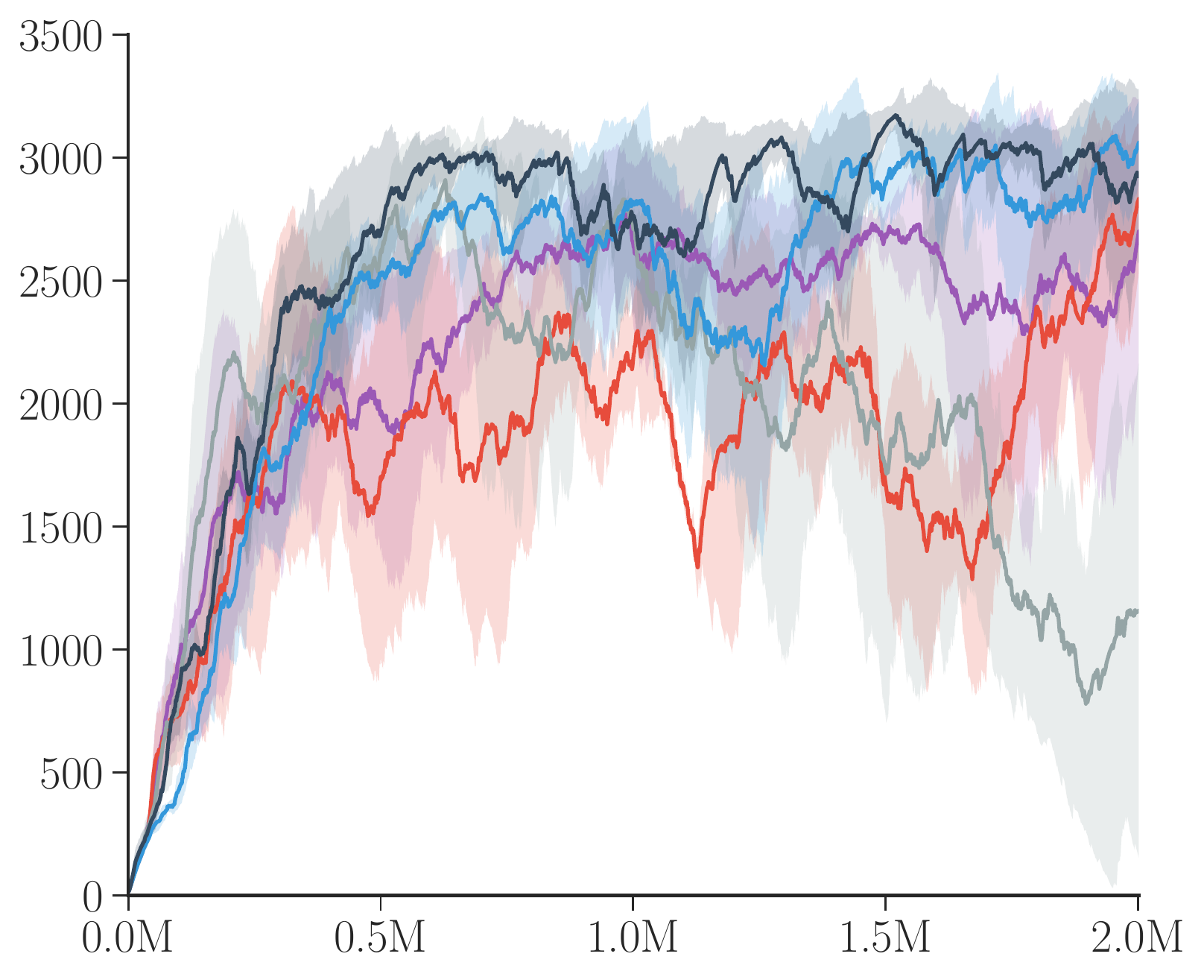} &
\includegraphics[width=.23\textwidth]{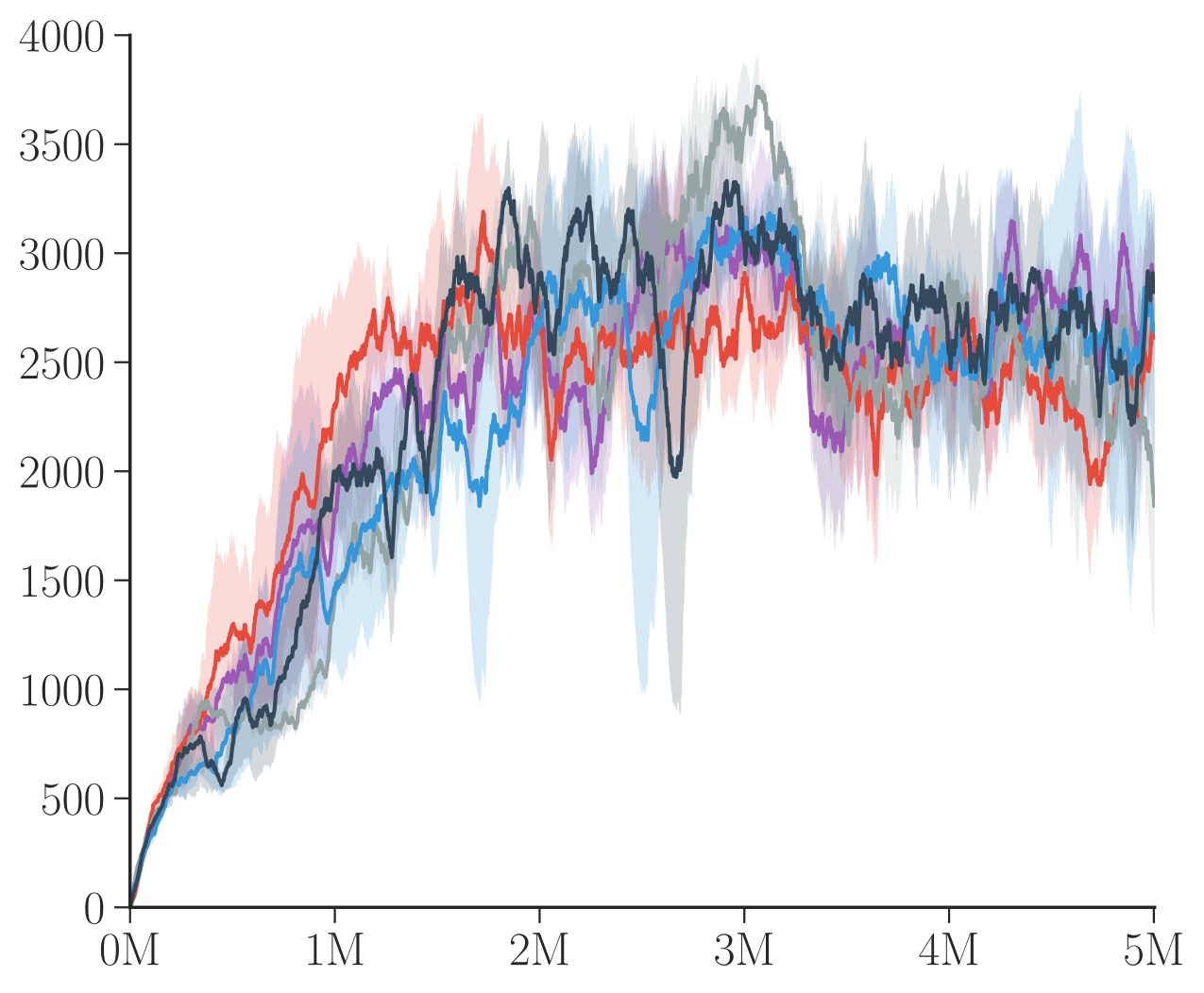} &
\includegraphics[width=.23\textwidth]{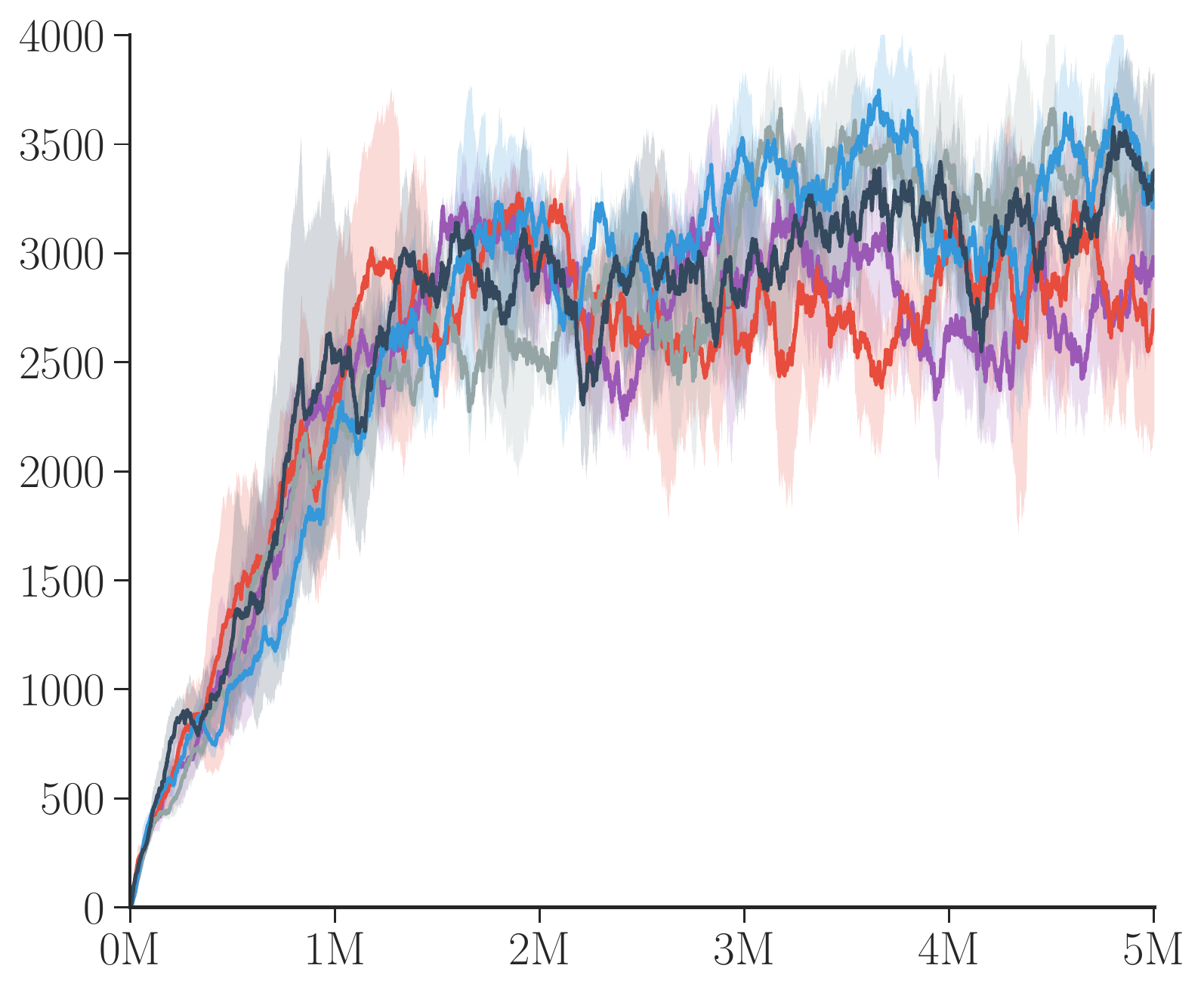} 
\llap{\raisebox{1.0em}{
\includegraphics[height=1.0cm]{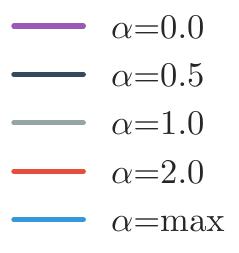}}} \\
\end{tabu}}
\vspace{-10bp}
\caption{Results on Hopper and Walker with different choices of $\alpha.$}
\label{fig:different_alpha}
\end{figure*}

In this section, we present the average reward of $\alpha$-divergences with
different choices of $\alpha$ on Hopper and Walker with both score-function and reparameterization gradient estimators. 
We can see from Figure~\ref{fig:different_alpha}
that $\alpha=0.5$ and $\alpha = +\infty$ (denoted by $\alpha=\max$ in the legends) perform consistently better
than standard KL divergence ($\alpha=0$), which is used the original SAC paper.

\subsection{Tail-adaptive $f$-divergence with score function estimation}

\begin{figure*}[ht]
\centering
{
\renewcommand{\arraystretch}{1} 
\begin{tabu}{ccccc}
\tiny{Ant} &  \tiny{HalfCheetah} &  \tiny{Humanoid(rllab)}\\
\includegraphics[width=.25\textwidth]{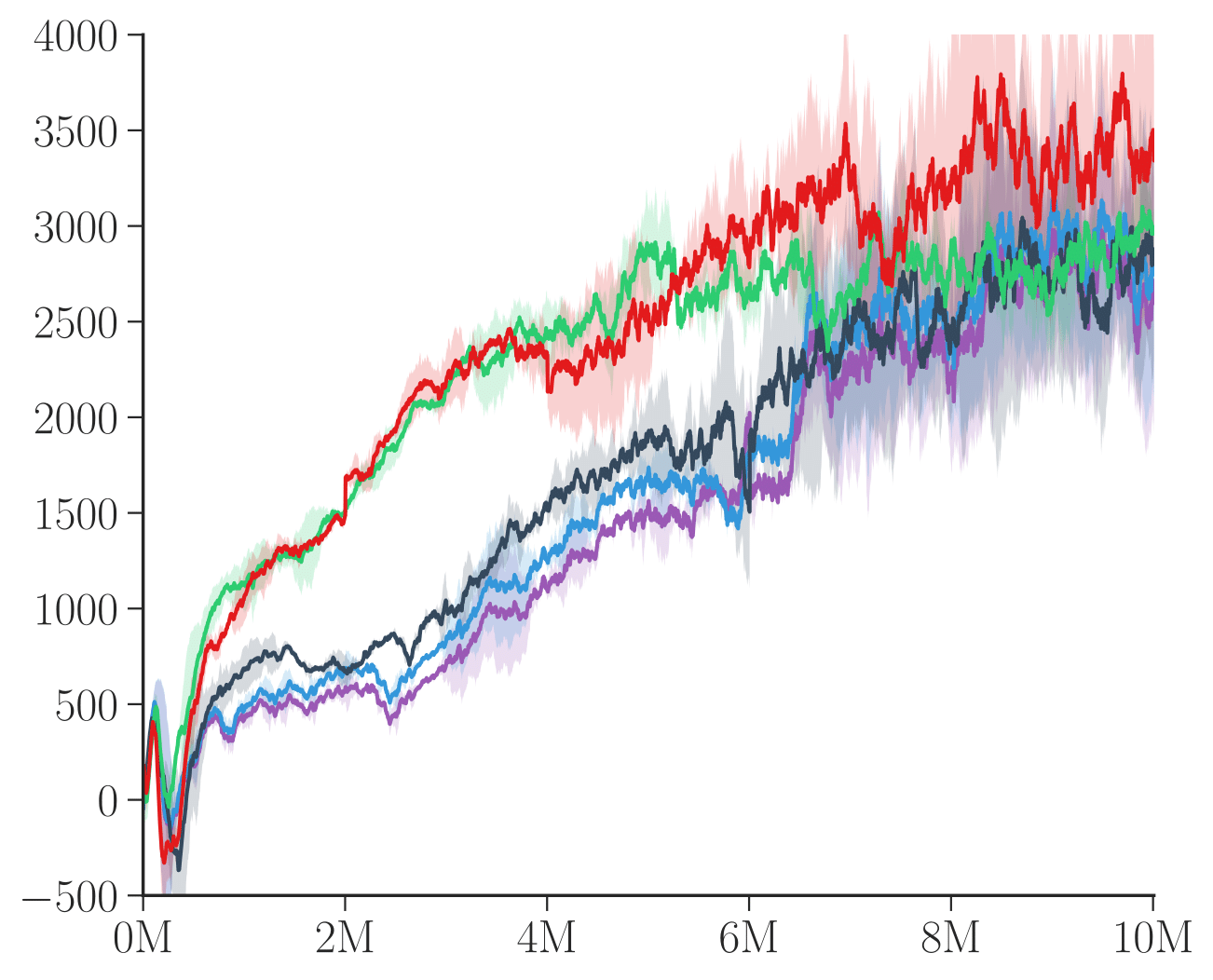}&
\includegraphics[width=.25\textwidth]{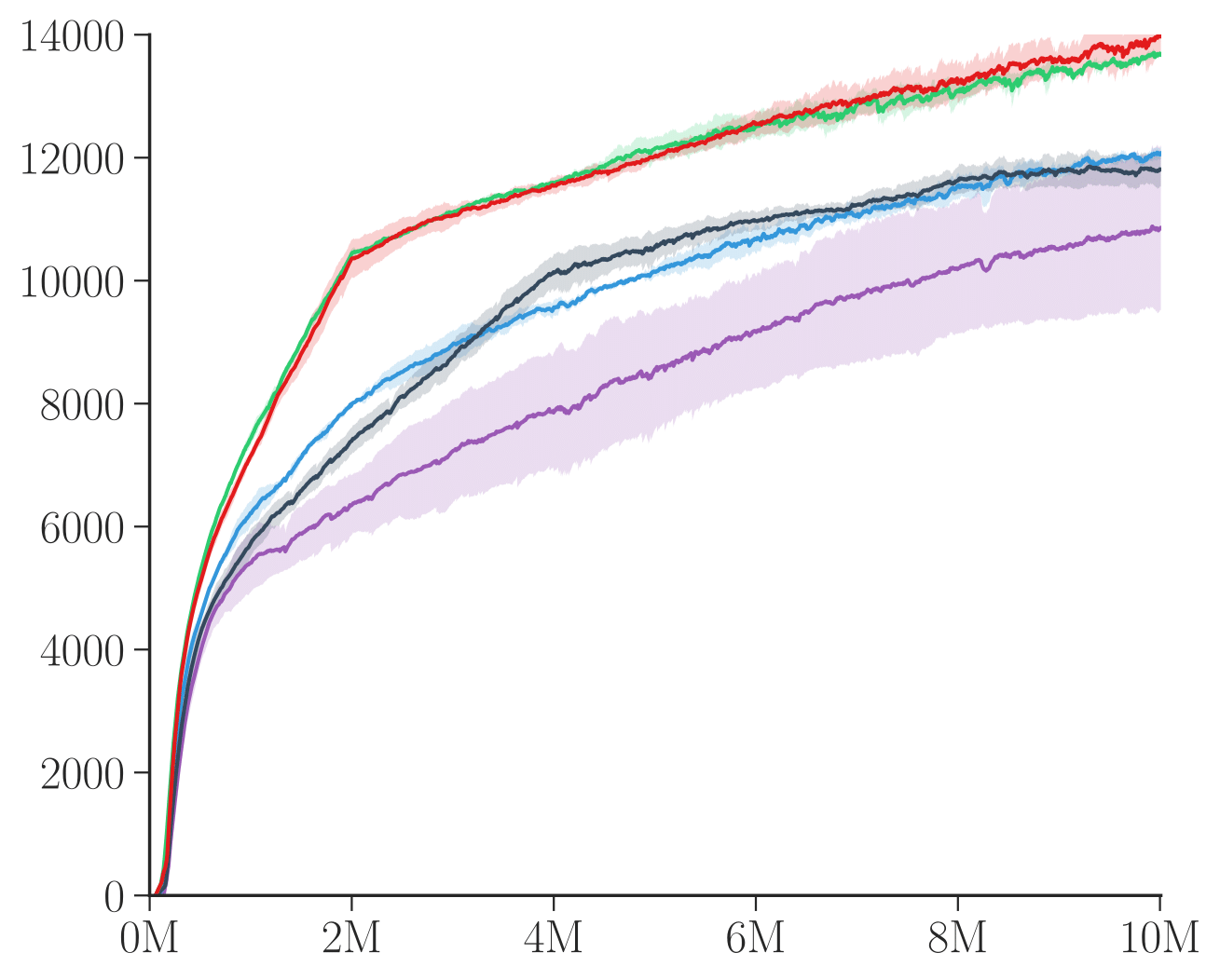}&
\includegraphics[width=.25\textwidth]{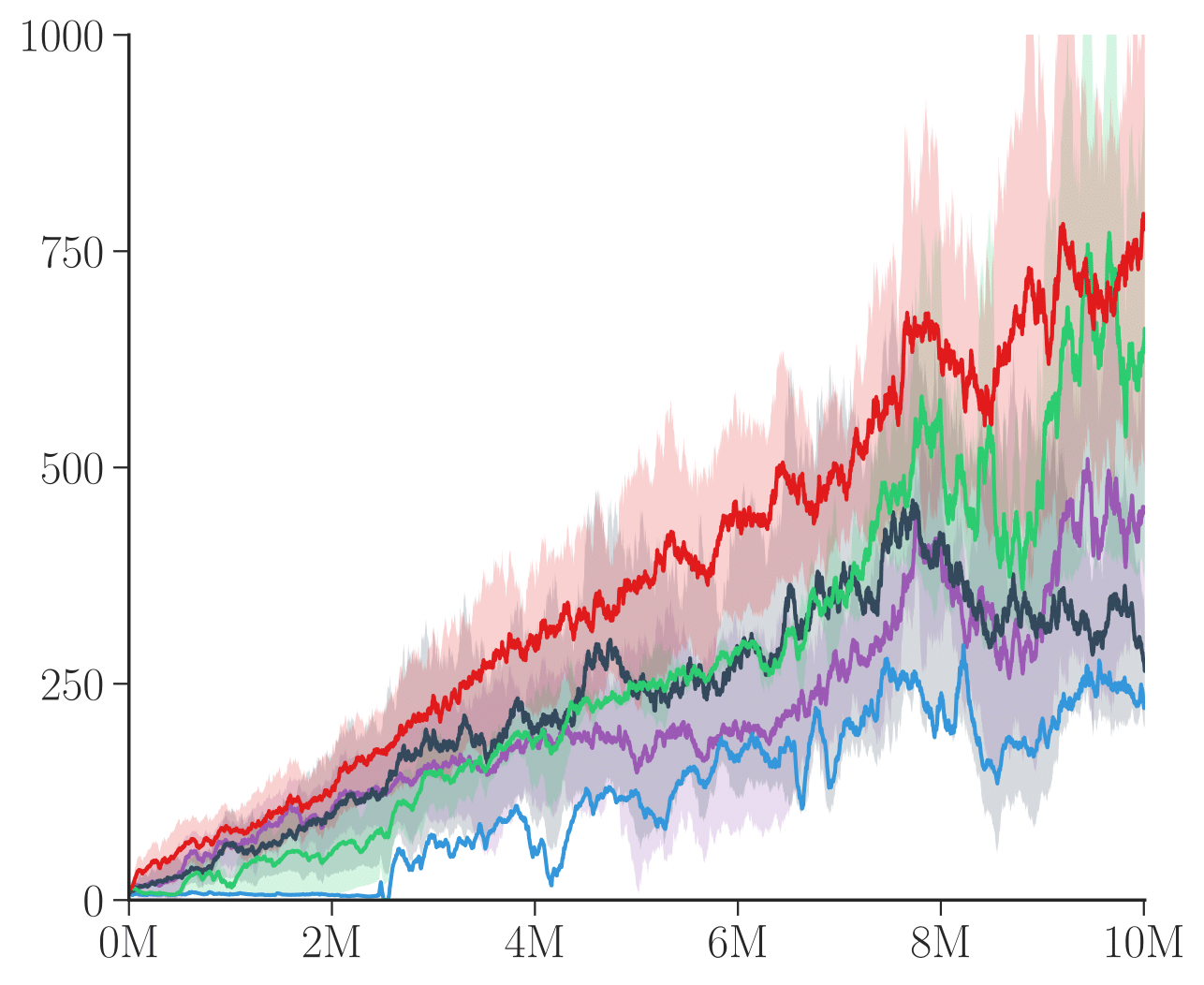}\\
\tiny{Walker} &  \tiny{Hopper} & \tiny{Swimmer(rllab)} \\
\includegraphics[width=.25\textwidth]{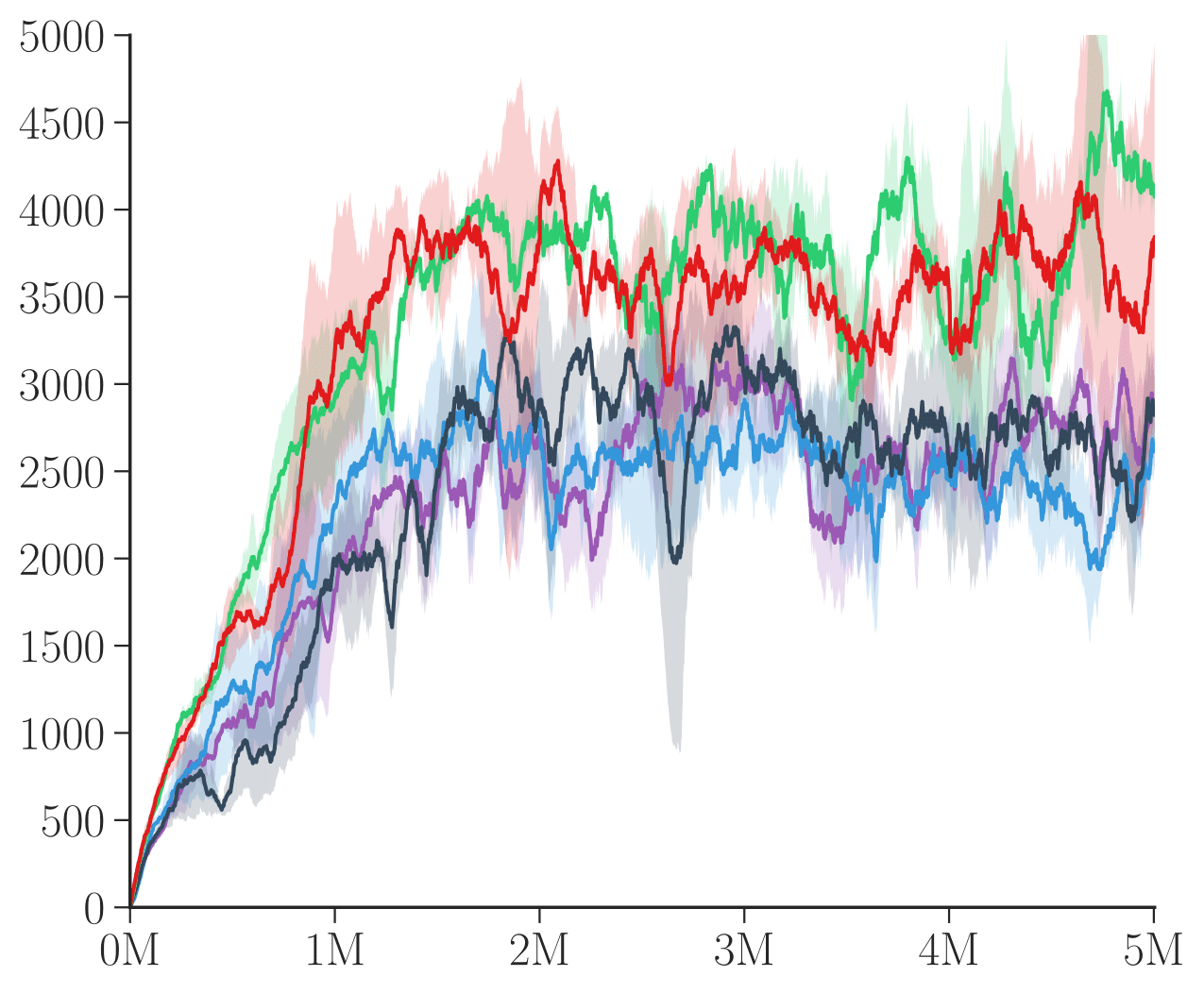} &
\includegraphics[width=.25\textwidth]{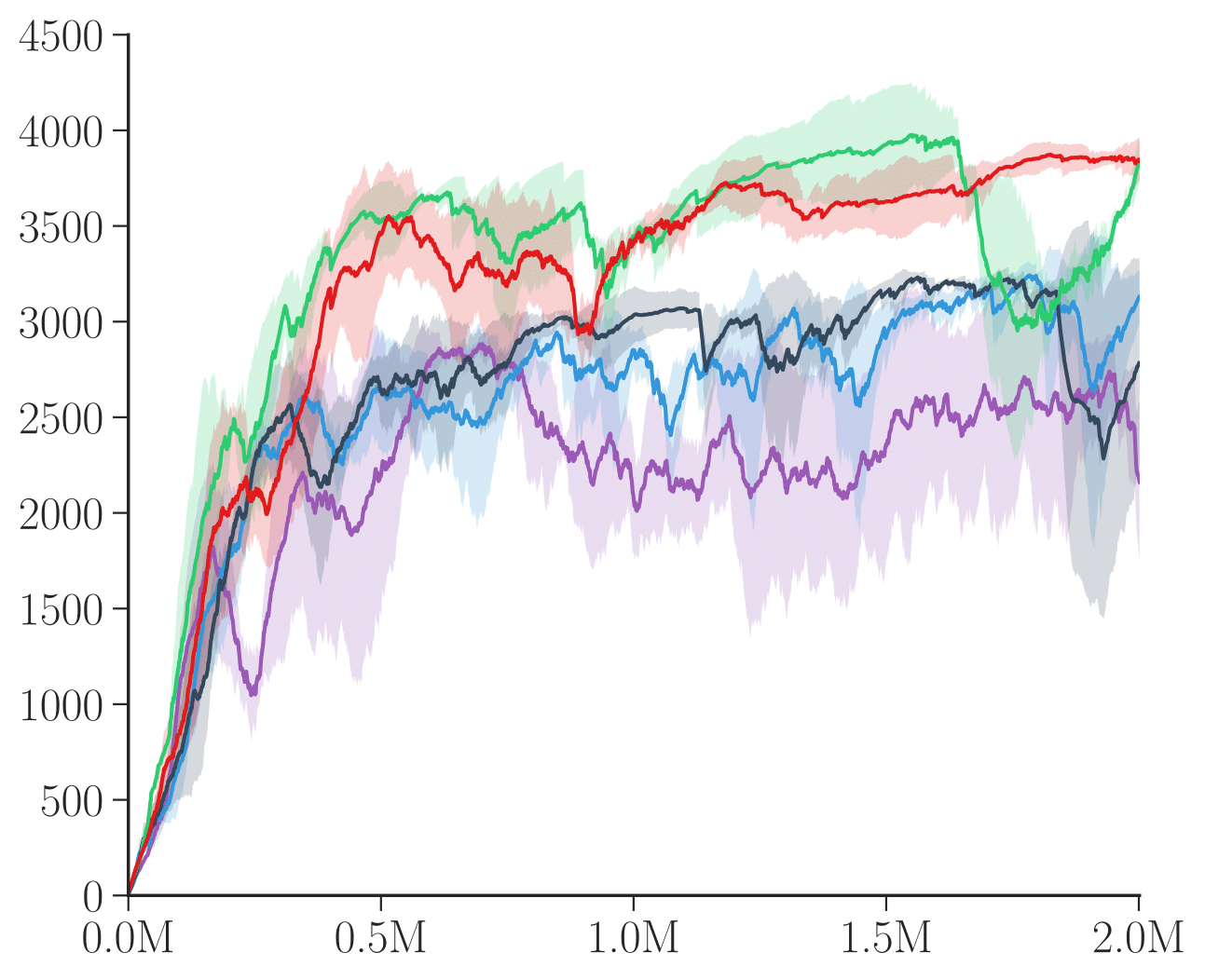} &
\includegraphics[width=.25\textwidth]{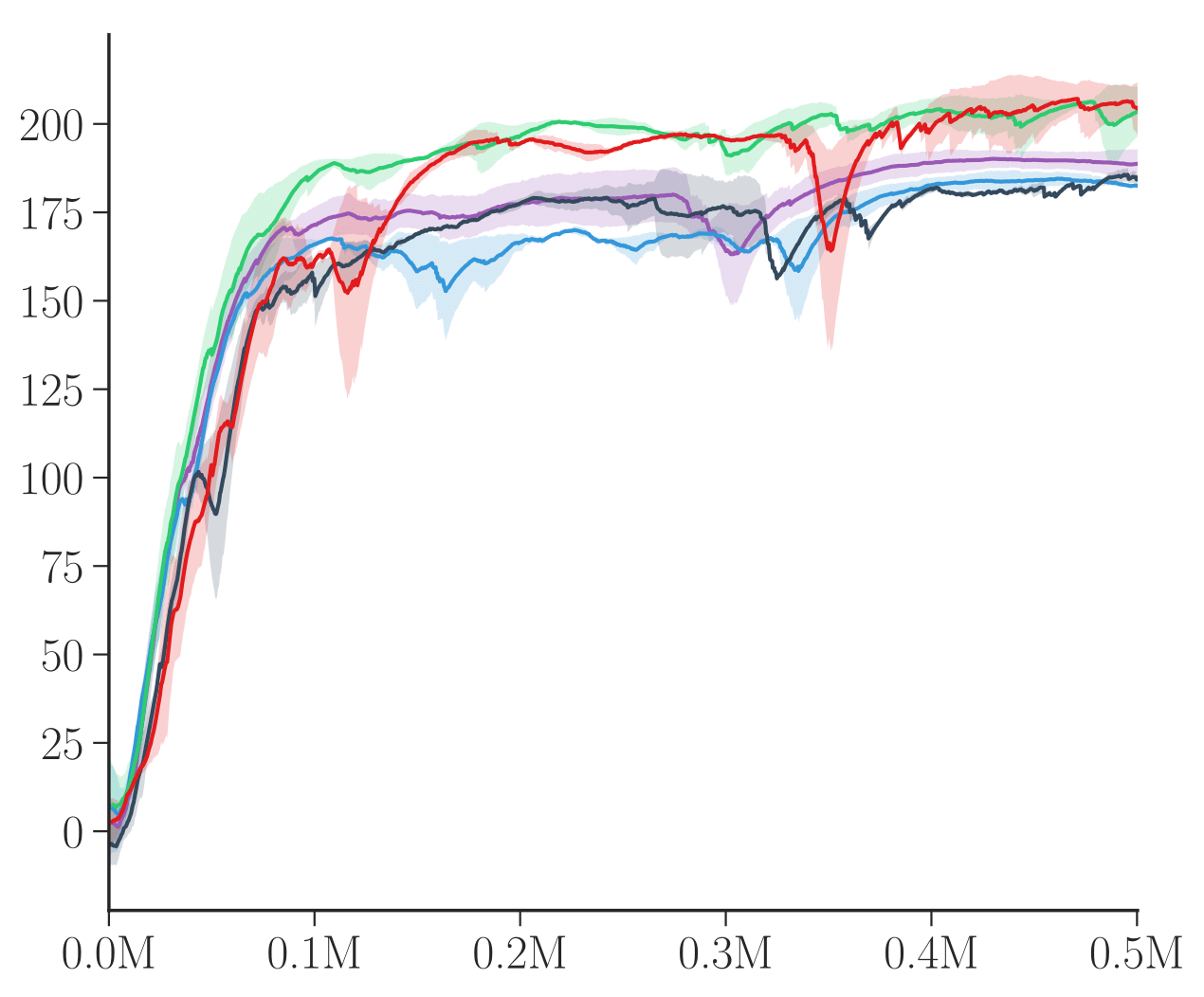}
\llap{\raisebox{1.2em}{
\includegraphics[height=1.2cm]{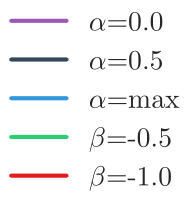}}} \\
\end{tabu}}
\vspace{-10bp}
\caption{Results of average rewards with score function gradients.}
\label{fig:score_different_f}
\end{figure*}

In this section, we  investigate optimization with score function gradient estimators (Algorithm~\ref{alg2}).  
The results in Figure~\ref{fig:score_different_f} show that  
our tail-adaptive $f$-divergence tends to yield better performance across all environments tested. 

